\documentclass[conference]{IEEEtran}
\usepackage{times}
\usepackage[dvipsnames]{xcolor}

\usepackage[numbers]{natbib}
\usepackage{multicol}
\usepackage{hyperref}
\usepackage{bm}
\usepackage[dvipsnames]{xcolor}
\hypersetup{
    colorlinks=true,
    linkcolor=blue,
    filecolor=magenta,      
    urlcolor=StanfordRed
    }

\usepackage{graphics} 
\usepackage{epsfig} 
\usepackage{mathptmx} 
\usepackage{times} 
\usepackage{amsmath} 
\usepackage{amssymb}  
\usepackage{siunitx} 
\usepackage{textcomp}
\usepackage{gensymb} 
\usepackage{booktabs}
\usepackage{multirow}
\usepackage{xcolor}
\usepackage{cleveref} 
\let\labelindent\relax
\usepackage{enumitem}
\usepackage{xspace}

\usepackage{makecell}
\usepackage{multicol}
\usepackage{rotating}
\usepackage{courier}

\usepackage{subcaption} 
\usepackage[font=small]{caption}

\usepackage{pifont}
\definecolor{MyDarkBlue}{rgb}{0,0.08,1}
\definecolor{airforceblue}{rgb}{0.36, 0.54, 0.66}
\definecolor{MyDarkGreen}{rgb}{0.02,0.6,0.02}
\definecolor{MyDarkRed}{rgb}{0.8,0.02,0.02}
\definecolor{MyDarkOrange}{rgb}{0.40,0.2,0.02}
\definecolor{MyPurple}{RGB}{111,0,255}
\definecolor{MyRed}{rgb}{1.0,0.0,0.0}
\definecolor{MyGold}{rgb}{0.75,0.6,0.12}
\definecolor{MyDarkgray}{rgb}{0.66, 0.66, 0.66}
\definecolor{MyPink}{rgb}{0.9, 0.33, 0.5}
\definecolor{MyCyan}{rgb}{0., 0.4, 0.4}
\definecolor{MyBlue}{rgb}{0.5, 0.8, 0.9}
\definecolor{AbsoluteColor}{rgb}{0.76, 0.2, 0.2}
\definecolor{DeltaColor}{rgb}{0.87, 0.72, 0.3}
\definecolor{RelativeColor}{rgb}{0.04, 0.33, 0.58}
\definecolor{StanfordRed}{rgb}{0.549, 0.082, 0.082}

\newcommand{\mypara}[1]{\par\vspace*{0mm} \textbf{\underline{{#1}}}}

\begin{document}


\title{\huge{UMI-3D: Extending Universal Manipulation Interface from Vision-Limited to 3D Spatial Perception}}


\author{\authorblockN{Ziming Wang$^{1, 2}$} 
${}^{1}$HKU, ${}^{2}$ USTC\\
\url{https://umi-3d.github.io}
}



%


\IEEEpeerreviewmaketitle
\twocolumn[{%
	\renewcommand\twocolumn[1][]{#1}%
	\maketitle
        \vspace{-4mm}
	\begin{center}
		\includegraphics[width=\textwidth]{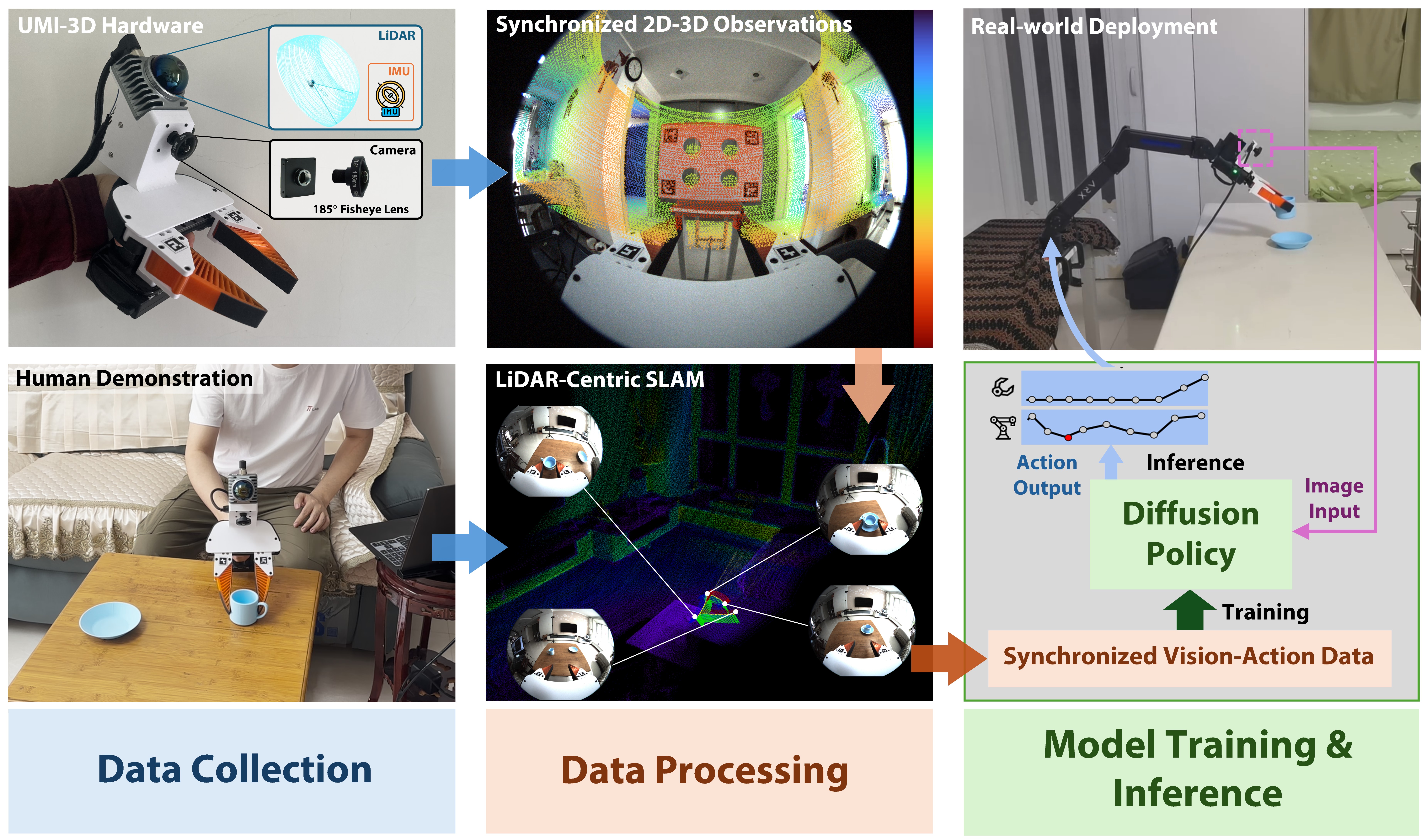}
		\captionof{figure}{ Overview of the UMI-3D system. From left to right, the pipeline consists of three stages: 
    (1) \textbf{Hardware and data collection}, where a wrist-mounted multimodal sensor suite (LiDAR, camera, and IMU) captures synchronized observations during human demonstrations; 
    (2) \textbf{LiDAR-centric SLAM and data processing}, which produces accurate metric-scale trajectories and consistent 3D representations; 
    (3) \textbf{Policy training and deployment}, where synchronized vision-action data are used to train visuomotor policies for real-world manipulation. }
  \label{fig:umi3d_overview}
	\end{center}
}]



\begin{abstract}
We present UMI-3D, a multimodal extension of the Universal Manipulation Interface (UMI) for robust and scalable data collection in embodied manipulation. While UMI enables portable, wrist-mounted data acquisition, its reliance on monocular visual SLAM makes it vulnerable to occlusions, dynamic scenes, and tracking failures, limiting its applicability in real-world environments.  UMI-3D addresses these limitations by introducing a lightweight and low-cost LiDAR sensor tightly integrated into the wrist-mounted interface, enabling LiDAR-centric SLAM with accurate metric-scale pose estimation under challenging conditions. We further develop a hardware-synchronized multimodal sensing pipeline and a unified spatiotemporal calibration framework that aligns visual observations with LiDAR point clouds, producing consistent 3D representations of demonstrations.  Despite maintaining the original 2D visuomotor policy formulation, UMI-3D significantly improves the quality and reliability of collected data, which directly translates into enhanced policy performance. Extensive real-world experiments demonstrate that UMI-3D not only achieves high success rates on standard manipulation tasks, but also enables learning of tasks that are challenging or infeasible for the original vision-only UMI setup, including large deformable object manipulation and articulated object operation.  The system supports an end-to-end pipeline for data acquisition, alignment, training, and deployment, while preserving the portability and accessibility of the original UMI. All hardware and software components are open-sourced to facilitate large-scale data collection and accelerate research in embodied intelligence: \href{https://umi-3d.github.io}{https://umi-3d.github.io}.
\end{abstract}

\section{Introduction}
Recent advances in data-driven robot learning have highlighted the importance of scalable and high-quality demonstration data \cite{chi2025diffusion,zhao2023learning, black2024pi_0}. The Universal Manipulation Interface (UMI) represents a significant step toward this goal by enabling low-cost, portable data collection through wrist-mounted sensing, effectively bridging the embodiment gap between human demonstrations and robot execution \cite{chi2024universal}. By leveraging visual perception and handheld operation, UMI allows users to collect diverse manipulation data in real-world environments without specialized infrastructure, opening a new paradigm for embodied data acquisition beyond traditional laboratory settings. 

Recent studies have further revealed that the effectiveness of such data collection paradigms is not merely empirical, but follows predictable scaling behavior. In particular, prior work on data scaling laws in robotic imitation learning demonstrates that policy performance improves consistently with increasing amounts of high-quality demonstration data, following power-law trends analogous to those observed in language and vision models \cite{hu2024data}. These findings indicate that the ability to collect large-scale, diverse, and well-aligned data is a primary bottleneck for advancing embodied intelligence. This perspective is further reinforced by recent large-scale efforts such as GEN-0, which trains generalist manipulation policies on over 270,000 hours of real-world interaction data, explicitly validating the scaling law hypothesis in robotics \cite{generalist2025gen0}. As demonstrated, increasing the scale of pretraining data leads to predictable improvements in downstream task performance and generalization, highlighting the critical role of scalable data acquisition systems in enabling next-generation embodied foundation models. Taken together, these results position systems like UMI not only as data collection tools, but as key infrastructure for scaling robot learning.

Despite its effectiveness as a scalable data collection interface, UMI still inherits fundamental limitations from its reliance on monocular visual SLAM for pose estimation \cite{campos2021orb}. As acknowledged in the original work, the system requires environments with sufficiently rich visual features to maintain stable tracking, and may fail in textureless regions such as blank walls or poorly lit scenes \cite{chi2024universal}. More critically, subsequent empirical studies on large-scale UMI data collection further reveal that visual SLAM is highly sensitive to occlusions and dynamic scene changes. For instance, during the manipulation of large objects that partially or fully block the camera view, such as opening doors or drawers, the SLAM system may misinterpret motion or even lose tracking entirely, thereby limiting the range of feasible tasks \cite{hu2024data}. Similarly, environments with insufficient visual features or the presence of moving distractors can lead to frequent tracking failures, requiring careful scene design or post hoc data filtering. These limitations introduce a fundamental tension in scaling up data collection. While large-scale and diverse data are essential for learning generalizable policies, the reliability of visual SLAM imposes implicit constraints on the environments, tasks, and interactions that can be captured. As a result, the data collection process becomes biased toward scenarios that are favorable for visual SLAM, hindering robustness and limiting the scalability of embodied learning systems, while also increasing the cost of data processing and curation. These limitations suggest that the bottleneck of scalable embodied data collection does not lie merely in interface design, but fundamentally in the reliability of pose estimation.

In this work, we propose UMI-3D, a LiDAR-centric extension of the UMI framework that rethinks the role of SLAM in embodied data acquisition. An overview of the proposed system is shown in Fig.~\ref{fig:umi3d_overview}. Instead of treating SLAM as an auxiliary component for tracking, we elevate SLAM to a core mechanism for ensuring metric-consistent, temporally aligned perception-action data.

Specifically, UMI-3D introduces a lightweight, wrist-mounted LiDAR-visual sensing system with tightly synchronized multimodal measurements, enabling robust and self-contained pose estimation without reliance on external infrastructure or environment-specific assumptions. By leveraging LiDAR-centric SLAM, our system remains stable under occlusions, textureless regions, illumination variations, and dynamic interactions, effectively overcoming the fundamental limitations of visual SLAM. This shift has two important consequences. First, it significantly improves the reliability and quality of collected data, reducing the need for environment curation and post hoc filtering. Second, and more importantly, it expands the feasible task distribution for data collection. Tasks that were previously challenging or infeasible under visual SLAM, such as curtain pulling, door opening, and manipulation under severe occlusion, become consistently collectable at scale.

We argue that this transition from vision-centric to geometry-aware perception is essential for scaling embodied data collection in real-world environments. In this view, SLAM serves not only as a localization module, but as a fundamental mechanism that aligns perception and action in a unified metric space, forming the foundation for scalable embodied learning systems.

\begin{figure*}[t] 
    \centering
    \includegraphics[width=\linewidth]{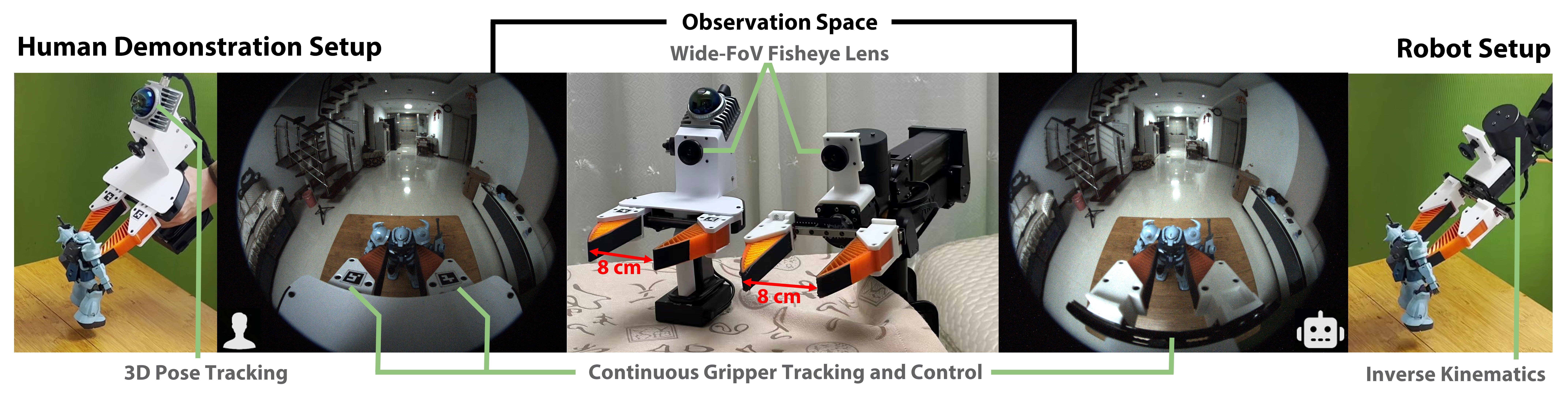}
    \caption{\textbf{UMI-3D Demonstration Interface Design.} 
    The system adopts a wrist-mounted sensing design that ensures consistent observation between human demonstrations (left) and robot execution (right). 
    A wide-FoV fisheye camera provides a shared observation space across embodiments, while continuous gripper tracking enables precise action recording and control. This design establishes a unified perception-action interface for data collection and policy deployment.}
    \label{fig:hardware}
    \vspace{-4mm}
\end{figure*}

\vspace{-2mm}
\section{Related Works} 

\subsection{UMI Variants and System Evolution}
The effectiveness of UMI-style systems critically depends on the mechanism used for end-effector pose estimation. As discussed above, the reliance on monocular visual SLAM introduces fundamental limitations in robustness and scalability, particularly under occlusions, dynamic interactions, and feature-sparse environments.  To address these challenges, recent works have explored alternative localization strategies for UMI-like systems, which can be broadly categorized into two directions.

The first direction replaces visual SLAM entirely with external tracking systems. For example, motion capture solutions based on infrared tracking, such as HTC Vive Tracker, provide millimeter-level accuracy and robust pose estimation in controlled environments \cite{liu2026rdt2}. However, these systems require external infrastructure such as base stations, which significantly reduces portability, increases cost, and contradicts the in-the-wild design philosophy of UMI. Furthermore, their performance may degrade in environments with strong lighting, occlusions, or limited installation space.

The second direction focuses on improving usability and robustness within the visual SLAM paradigm. A representative example is FastUMI \cite{liu2024fastumi}, which replaces the original VIO pipeline with an off-the-shelf tracking module, such as the Intel RealSense T265, thereby reducing calibration complexity and enabling plug-and-play deployment.  Similarly, other systems leverage consumer devices with built-in SLAM capabilities, such as Meta Quest headsets \cite{xu2025exumi,zeng2025activeumi} or iPhone ARKit \cite{xu2025dexumi, ha2024umi, choi2026wild, fang2026robopocket, xu2026hommi}, to provide integrated pose tracking without requiring custom SLAM implementations. These approaches lower the barrier for non-expert users in data collection.

However, we argue that these approaches primarily improve engineering convenience rather than addressing the intrinsic limitations of visual SLAM. First, commercial SLAM modules are typically closed-source and operate as black-box systems, limiting their adaptability to different tasks, environments, and sensing configurations. This restricts the ability to leverage the rich body of research in the SLAM community for task-specific optimization. Second, decoupling the tracking module from the observation sensor introduces temporal inconsistency. In systems such as FastUMI \cite{liu2024fastumi, huang2025umigen}, pose estimation is performed by a dedicated tracking device (e.g., T265), while visual observations are captured by a separate camera (e.g., GoPro). This disrupts the temporal alignment between observations and actions, potentially introducing systematic errors in dynamic manipulation tasks. In contrast, the original UMI design uses a unified sensing source for both observation and SLAM, ensuring strict temporal alignment. Finally, the limitations of visual SLAM are fundamentally rooted in its reliance on image features. While commercial systems may improve robustness through better engineering and optimization, they cannot fully overcome challenges such as textureless environments, severe occlusions, or dynamic disturbances.

These observations suggest that improving visual SLAM alone is insufficient for achieving reliable and scalable in-the-wild manipulation data collection. Existing UMI variants primarily focus on improving usability or robustness within specific implementations, yet lack a principled understanding of what constitutes a reliable pose estimation mechanism for embodied data collection.

To bridge this gap, we revisit the design of UMI-style systems from a principled perspective. Building upon the discussion in Sec.~I, we identify three key principles that a well-designed UMI system should satisfy:
\begin{enumerate}[leftmargin=4mm]
    \item \textbf{Self-contained pose estimation.} The system should rely exclusively on sensors mounted on the gripper, eliminating the need for external infrastructure and enabling fully portable deployment.
    
    \item \textbf{Spatiotemporal consistency in multimodal sensing.} Multimodal perception must maintain strict alignment through accurate time synchronization and precise spatial calibration, ensuring coherent cross-sensor observations.
    
    \item \textbf{Consistency of localization accuracy.} 
    A reliable pose estimation system should maintain stable and consistent accuracy across diverse environments, tasks, and interaction conditions, rather than exhibiting strong performance only under favorable or constrained scenarios. This consistency is critical for ensuring reliable data collection at scale, where environmental conditions and task dynamics cannot be controlled.
\end{enumerate}

These principles provide a unified framework for evaluating existing UMI variants and directly guide the design of our proposed system. In particular, they motivate a shift from vision-centric localization toward LiDAR-centric perception, which forms the foundation of our approach.

\subsection{From Visual SLAM to LiDAR-centric SLAM}
SLAM systems can be broadly categorized into vision-based and LiDAR-based paradigms, which have evolved largely in parallel over the past decades. While visual SLAM benefits from low-cost sensing and rich semantic information, LiDAR SLAM provides superior geometric accuracy and robustness in challenging environments. As a result, the two paradigms have each established themselves as dominant approaches across different application domains.

Vision-based SLAM has undergone substantial development over the past decades, evolving along both methodological and system-level dimensions. From a methodological perspective, existing approaches can be broadly categorized into feature-based \cite{mur2015orb, campos2021orb, qin2018vins}, direct \cite{engel2014lsd}, and hybrid \cite{forster2014svo} methods, which differ in whether geometric correspondences or photometric consistency are exploited for motion and structure estimation. 

At the system level, modern visual SLAM frameworks have progressively advanced from frame-to-frame odometry to keyframe-based optimization, and further to globally consistent mapping through loop closure and long-term data association \cite{mur2015orb, campos2021orb}. This evolution has significantly improved accuracy and robustness, enabling reliable operation across a wide range of scenarios. To address the inherent limitations of pure vision, such as scale ambiguity and sensitivity to visual degradation, visual-inertial systems have been introduced to incorporate inertial measurements within tightly-coupled optimization frameworks, further enhancing motion estimation accuracy and robustness \cite{qin2018vins}.

Despite these advances, visual SLAM remains fundamentally constrained by its reliance on image formation. Its performance degrades under challenging visual conditions such as illumination changes, motion blur, and texture scarcity \cite{qin2018vins}, and it typically lacks direct geometric observability. Moreover, the implicit assumption of photometric consistency and predominantly static environments further limits its applicability in dynamic and complex real-world scenarios. These inherent limitations motivate the development of LiDAR-centric SLAM systems, which provide direct and reliable geometric measurements and exhibit superior robustness in perceptually degraded and dynamically evolving environments \cite{xu2022fast}.

In contrast to vision-based approaches, LiDAR-centric SLAM directly leverages active depth measurements, providing accurate and illumination-invariant geometric observations \cite{zhang2014loam}. Early LiDAR SLAM systems established a structured pipeline that decouples high-frequency odometry from low-frequency mapping, typically relying on geometric feature extraction such as edge and planar primitives for scan registration \cite{zhang2014loam, shan2018lego}. This design enables real-time performance while maintaining low drift through hierarchical optimization. Building upon this foundation, subsequent works have focused on improving system efficiency and robustness. LiDAR-inertial fusion frameworks introduce tightly-coupled state estimation to compensate for motion distortion and enhance pose accuracy \cite{shan2020lio, xu2022fast}, while recent methods further eliminate hand-crafted feature extraction by directly registering raw point clouds to maps, improving adaptability across different sensing modalities and environments \cite{xu2022fast}. Beyond system design, recent advances have also revisited map representation. Instead of relying on raw point cloud maps, which are often computationally expensive and lack uncertainty modeling, more sophisticated representations such as voxelized and probabilistic maps have been proposed to enable efficient map updates, robust scan matching, and principled handling of measurement uncertainty \cite{yuan2022efficient}.

Overall, LiDAR SLAM systems provide direct geometric observability, strong robustness to perceptual degradation, and improved reliability in large-scale and dynamic environments. However, despite these advantages, the widespread adoption of LiDAR in robotic systems has historically been limited by its high cost, large form factor, and significant power consumption compared to vision sensors \cite{geiger2013vision, wang2023ustc}. Recent advances in sensor design and manufacturing have substantially reduced these barriers, enabling compact, low-cost, and energy-efficient LiDAR systems \cite{liu2021low, ren2025survey}. This technological shift not only narrows the gap between LiDAR and vision in terms of deployment constraints, making LiDAR a competitive and complementary alternative to vision-based approaches, but also enables new system designs on lightweight robotic platforms. In particular, it becomes feasible to integrate LiDAR directly at the end-effector, as exemplified by UMI-3D, which enables geometric perception at the point of interaction.

\vspace{-1mm}
\section{Method}
The design objectives of UMI-3D are aligned with those of the original Universal Manipulation Interface (UMI), aiming to provide a portable, information-rich, and reproducible framework for data collection and policy learning. Building upon the core principles of UMI, UMI-3D preserves its essential strengths while introducing key enhancements to address the limitations of visual SLAM, expand action diversity, and enable robust 3D spatial perception. In the following, we present the design of UMI-3D from three complementary perspectives: hardware design, data acquisition and processing, and policy interface, highlighting both inherited components and novel contributions.

\subsection{Demonstration Interface Design}
\label{sec:hardware_interface}

UMI-3D adopts the same mechanical design as UMI for data collection, retaining the trigger-activated, handheld 3D-printed parallel-jaw gripper with soft fingers. In addition to the top cover assembly for sensors mounting, its parts are completely interchangeable with the original UMI. This preserves the key advantages of portability, low cost, and reproducibility. The key distinction of UMI-3D lies in its sensing paradigm. Instead of relying solely on a wrist-mounted monocular RGB camera (e.g., GoPro in UMI), we integrate a hardware-synchronized multimodal sensor suite consisting of a LiDAR and a camera, while maintaining a similar sensor placement to preserve observation geometry and embodiment alignment as much as possible \hyperref[HD1]{HD1}.

Following the original UMI formulation, the central question becomes:
\begin{center}
\emph{How can we capture sufficient multimodal and spatial information 
for a wide variety of manipulation tasks using a compact, wrist-mounted sensor suite?}
\end{center}
To address this question, we design a wrist-mounted multimodal sensing interface that unifies observation and action across human demonstrations and robot execution.  This design establishes a consistent perception-action loop, where sensing remains tightly coupled with manipulation throughout data collection and deployment.  The overall demonstration setup and observation geometry are illustrated in Fig.~\ref{fig:hardware}.

\textbf{HD1. Wrist-mounted multimodal sensors as primary observation interface.}  
\label{HD1}
A key design principle of UMI-3D is to co-locate all sensing modalities directly on the gripper, ensuring that perception remains tightly coupled with manipulation. To this end, sensors are rigidly attached to the gripper and move together with it across both human demonstrations and robot execution, as illustrated in Fig.~\ref{fig:hardware}. This configuration ensures consistent observation geometry and enables seamless transfer between data collection and deployment.

Within this unified interface, the camera and LiDAR serve complementary but distinct roles. The camera captures rich visual context for policy learning, providing appearance information critical for action inference, while the LiDAR, together with LiDAR-based SLAM, provides accurate 3D geometric structure and consistent SE(3) motion estimation for spatial understanding. This division of functionality enables a unified perception interface that jointly captures appearance, geometry, and motion.

In practice, this design is implemented using an industrial CMOS camera (Hikrobot MV-CB013-A0UC-S) instead of the consumer-grade GoPro used in UMI, offering improved system integration, lower latency, and greater flexibility for synchronization and secondary development, while preserving the camera-centric interface for policy learning.

This design retains and extends the key advantages of wrist-mounted sensing in UMI:
\begin{enumerate}[leftmargin=4mm]
    \item \textbf{Consistent cross-embodiment observation.} Wrist-mounted sensing ensures aligned viewpoints between demonstration and deployment.
    \item \textbf{Infrastructure-free deployment.} The system supports portable operation with both camera-only and multimodal configurations.
    \item \textbf{Motion-induced data diversity.} Dynamic viewpoints improve robustness by encouraging focus on task-relevant regions.
    \item \textbf{Multimodal geometric perception.} LiDAR introduces a parallel pipeline for reliable 3D structure and motion estimation \hyperref[HD3]{HD3}.
\end{enumerate}

\textbf{HD2. Wide-FoV sensing for sufficient visual context.}  
\label{HD2}
A key requirement for wrist-mounted perception is to capture sufficient visual context despite rapid motion and continuously changing viewpoints. Following UMI, which demonstrates the effectiveness of wide-FoV fisheye imaging for maintaining task-relevant observations during manipulation~\cite{chi2024universal,xue2026rethinking}, we adopt a similar design. In UMI-3D, we further extend this approach by using a wider FoV fisheye camera (approximately 185°) through an M12 lens system, expanding the observable region (see Fig.~\ref{fig:hardware}). As in UMI, we directly use raw fisheye images without undistortion, preserving central resolution while compactly encoding peripheral context. Beyond improving visual coverage, the wide-FoV design also enables substantial overlap with the LiDAR field of view, providing a foundation for consistent multimodal perception by ensuring that visual and geometric observations correspond to overlapping regions of the workspace. However, wide-FoV imaging alone does not provide explicit and reliable depth or motion estimation, motivating the use of direct geometric sensing.

\textbf{HD3. Consistent and geometry-grounded spatial perception.} 
\label{HD3}
In UMI, depth perception is approximated using a pair of side mirrors that create implicit stereo views within a single image. In contrast, UMI-3D replaces this indirect approximation with direct geometric sensing using LiDAR, providing explicit metric-scale depth and spatial structure. Beyond improving geometric perception, this transition also enables robust and drift-resistant state estimation through LiDAR-based SLAM, producing consistent SE(3) motion estimates alongside 3D structure. This significantly enhances the reliability of spatial understanding, particularly in challenging conditions such as low-texture regions, occlusions, and dynamic scenes. Importantly, this design does not alter the core camera-centric interface for policy learning; instead, LiDAR acts as a complementary modality that augments the visual observation space with reliable geometric and motion information. The following section describes how these multimodal signals are temporally synchronized and spatially aligned to form a coherent perception pipeline: \hyperref[SP1]{SP1},\hyperref[SP2]{SP2},\hyperref[SP3]{SP3}.

\textbf{HD4. Continuous gripper control.}  
\label{HD4}
Consistent with UMI, we adopt continuous gripper width control instead of binary open-close actions. The gripper width is tracked via ArUco markers \cite{garrido2014automatic}, and the system can implicitly record and regulate grasp forces through the deformation of soft fingers.

\textbf{Putting everything together.} The UMI-3D gripper weighs 1120g, with an external dimension of $L285mm \times W175mm \times H300mm$ and finger stroke of $80mm$. The 3D printed gripper has a BoM cost of \textbf{\$50}, while the Livox MID-360 LiDAR and Hikrobot camera total \textbf{\$650}. As shown in Fig. \ref{fig:hardware}, we can equip any robot arms with a compatible gripper and camera setup.

\subsection{SLAM and Data Processing Pipeline}
\label{sec:slam_pipeline}
To enable reliable multimodal perception and consistent policy training, we construct a unified SLAM and data processing pipeline that transforms raw sensor streams into temporally aligned, spatially calibrated, and geometrically consistent representations. This pipeline integrates hardware-level synchronization, multi-sensor calibration, LiDAR-inertial odometry, and structured data packaging, ensuring that visual observations, geometric measurements, and motion estimates are coherently aligned within a common reference frame. Such consistency is critical for bridging demonstration and deployment, as it provides accurate SE(3) state estimation and high-quality 3D spatial information while preserving compatibility with the camera-centric policy interface. We organize this pipeline into four components.

\begin{figure}[t]
    \centering
    \includegraphics[width=0.95\linewidth]{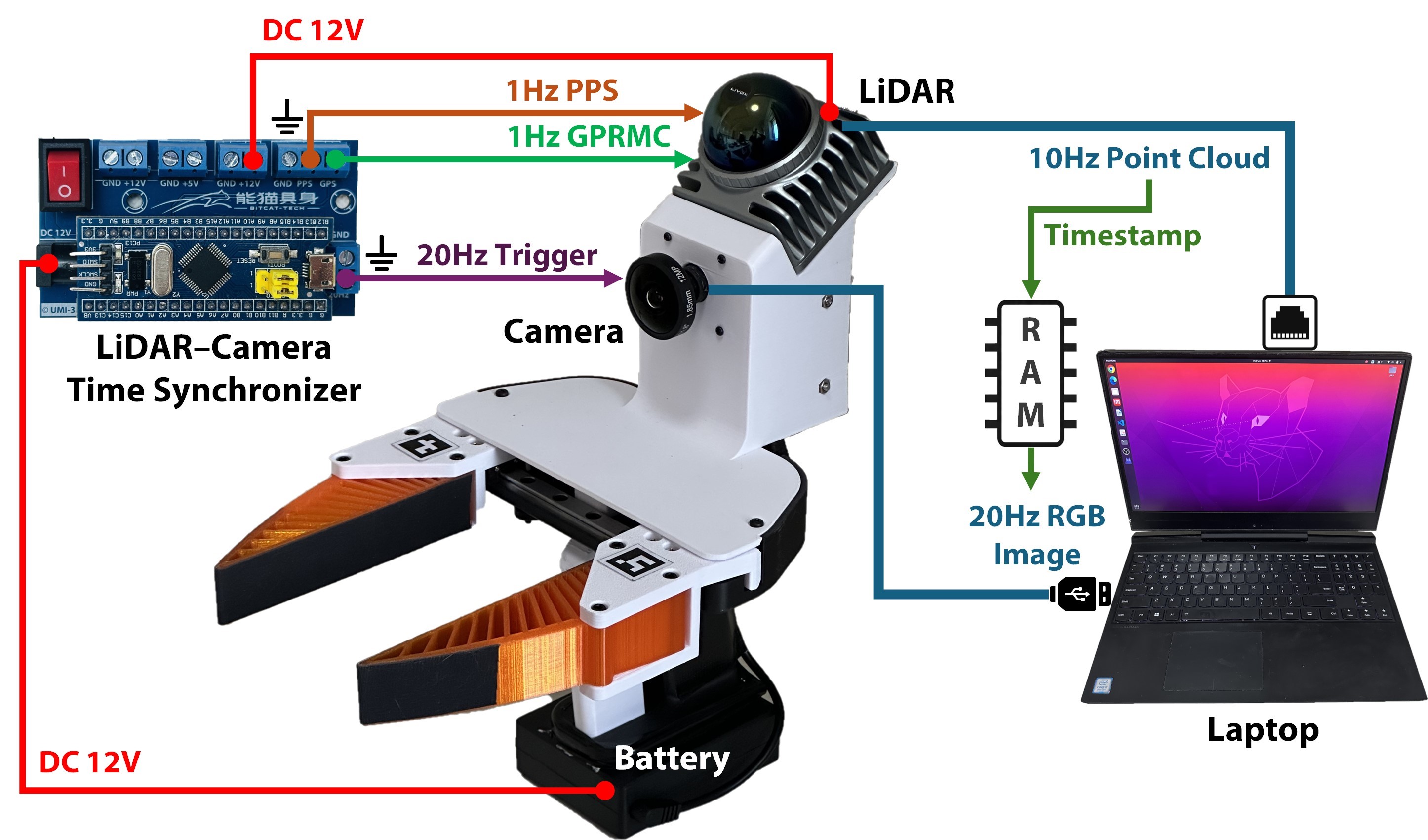}
    \caption{
    \textbf{Hardware-level temporal synchronization in UMI-3D.} 
    An STM32 microcontroller serves as a unified clock source for both LiDAR and camera. It provides a 1\,Hz PPS signal and a pseudo-GPRMC message to synchronize the LiDAR, while generating a 20\,Hz hardware trigger for the camera via frequency division. The LiDAR streams 10\,Hz point clouds over Ethernet, and the camera captures 20\,Hz RGB images via hardware triggering. Timestamps are shared between drivers to ensure consistent temporal alignment, and all data are recorded in a unified \texttt{rosbag} format.
    }
    \vspace{-5mm}
    \label{fig:time_sync}
\end{figure}

\textbf{SP1. Temporal synchronization.}
\label{SP1}
To ensure temporally consistent multimodal observations, we design a hardware-level synchronization mechanism that enforces a unified time base across LiDAR and camera sensors (Fig.~\ref{fig:time_sync}). Specifically, an STM32 microcontroller is employed as the central clock source, generating a 1\,Hz pulse-per-second (PPS) signal and a corresponding pseudo-GPRMC message to synchronize the LiDAR, while simultaneously producing a 20\,Hz trigger signal for the camera via frequency division. During data acquisition, the LiDAR streams point clouds at 10\,Hz over Ethernet, and the camera captures RGB images at 20\,Hz via hardware triggering. To achieve consistent timestamp alignment, the LiDAR ROS driver publishes timestamps to a shared memory buffer, from which the camera driver retrieves synchronized timestamps as its temporal reference. As a result, both visual and geometric observations are aligned under a common time base and recorded in a unified \texttt{rosbag} format. This hardware-synchronized design avoids drift and latency introduced by software-based synchronization, ensuring accurate temporal correspondence between image frames and point clouds, which is critical for downstream LiDAR-visual fusion and state estimation.

\textbf{SP2. Sensors calibration}
\label{SP2}

Accurate multimodal perception in UMI-3D requires precise calibration of both the fisheye camera and its geometric relationship with the LiDAR. We therefore adopt a two-stage calibration pipeline, including intrinsic calibration of the fisheye camera and extrinsic calibration between LiDAR and camera.

\textbf{SP2.1) Fisheye camera intrinsic calibration.}
We model the camera using the equidistant projection model, which is well-suited for ultra-wide-angle fisheye lenses. Given a 3D point $\mathbf{X}_c = (X, Y, Z)$ in the camera coordinate frame, its projection onto the image plane is defined as
\begin{equation}
    \theta = \arctan\left(\frac{\sqrt{X^2 + Y^2}}{Z}\right), \quad
    r = f \cdot \theta,
\end{equation}
where $\theta$ denotes the incident angle and $r$ is the radial distance from the principal point. Lens distortion is further modeled as a polynomial function of $\theta$.

To estimate the intrinsic parameters, we capture multiple views of a planar checkerboard calibration target under diverse orientations and distances, as shown in Fig.~\ref{fig:fisheye_calib}. The parameters, including focal length, principal point, and distortion coefficients, are obtained by minimizing the reprojection error:
\begin{equation}
    \min_{\mathbf{K}, \mathbf{D}} \sum_{i} \left\| \mathbf{u}_i - \pi(\mathbf{X}_i; \mathbf{K}, \mathbf{D}) \right\|^2,
\end{equation}
where $\pi(\cdot)$ denotes the equidistant projection function.

Accurate intrinsic calibration is essential for subsequent processing, including fisheye image undistortion, ArUco marker detection, and spatially consistent image cropping, all of which directly affect downstream multimodal alignment and policy learning.

\begin{figure}[!t]
    \centering
    \includegraphics[width=0.98\linewidth]{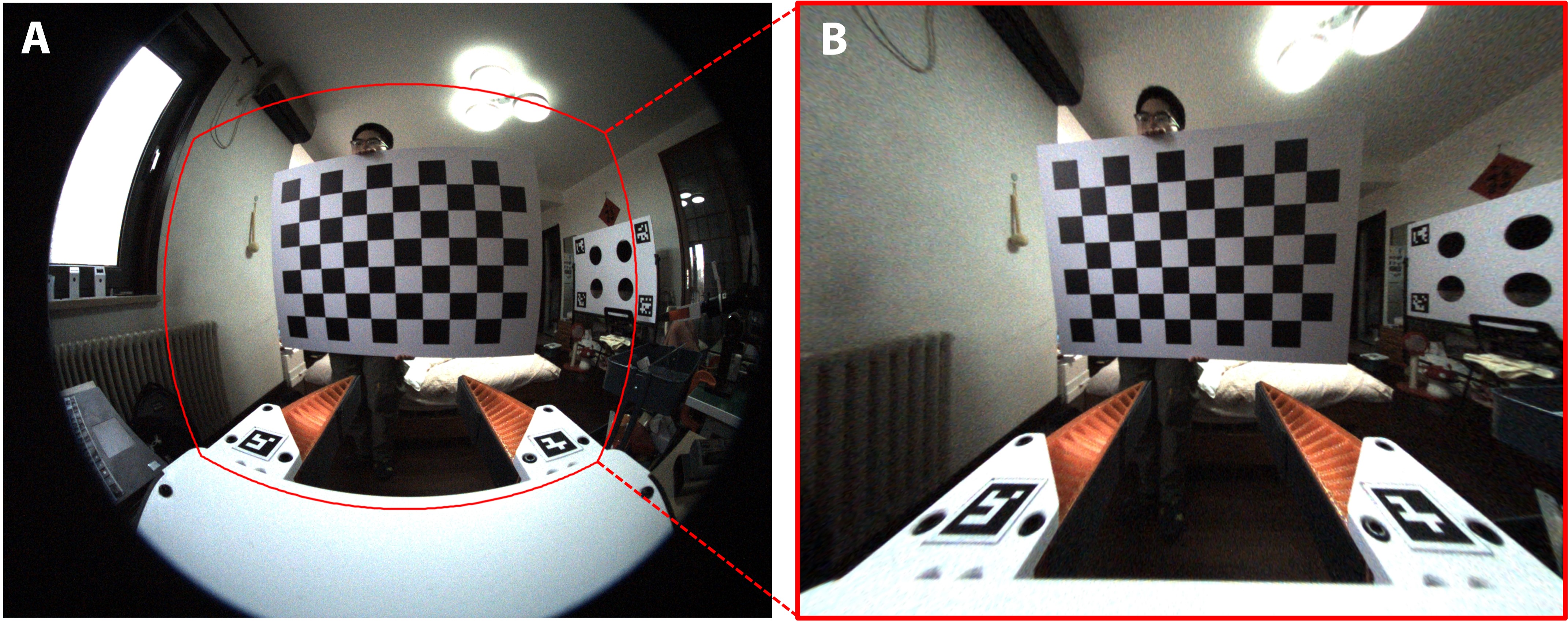}
    \caption{\textbf{Fisheye camera intrinsic calibration.} 
    (A) Raw fisheye image capturing a planar checkerboard calibration target under a wide field of view. 
    (B) Undistorted image using the estimated intrinsic parameters under the equidistant projection model.  
    This calibration enables precise pixel-to-ray mapping, which is essential for subsequent perception tasks including marker detection, spatial alignment, and multimodal fusion.} 
    \vspace{-7mm}
    \label{fig:fisheye}
\end{figure}

\begin{figure*}[!t]
    \centering
    \includegraphics[width=\linewidth]{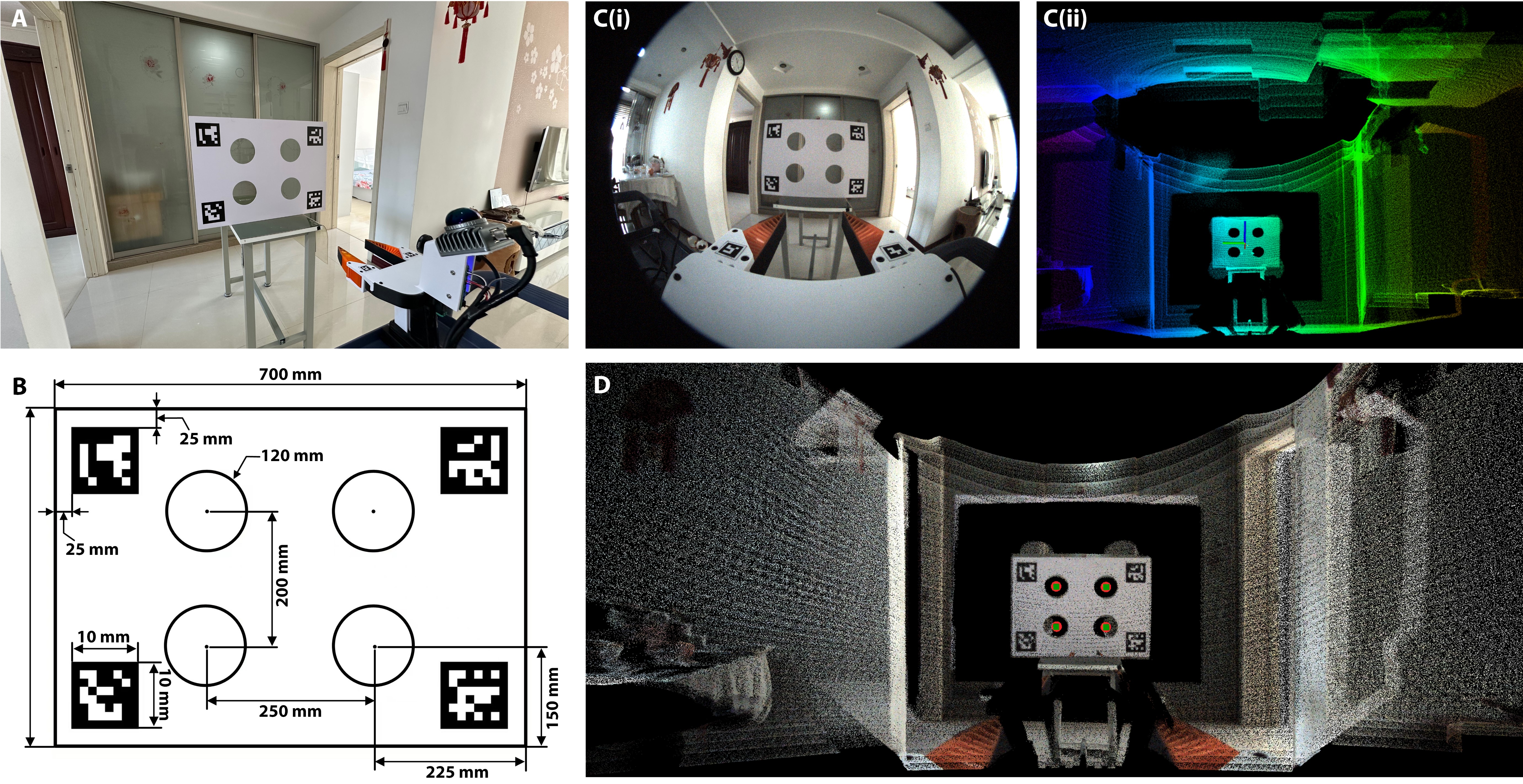}
   \caption{
    \textbf{LiDAR–camera extrinsic calibration.}
    (A) Calibration setup in a typical home environment, demonstrating the practicality and ease of deployment only with a calibration board. 
    (B) Design specification of the calibration target, including geometric layout and fiducial markers. 
    (C) Multimodal observations: (i) fisheye image with detected markers and calibration features; (ii) corresponding LiDAR point cloud. 
    (D) Colorized point cloud after applying the estimated extrinsic transformation, where LiDAR points are assigned RGB values by sampling from the corresponding image.
    This calibration enables accurate spatial alignment between visual and geometric observations, forming the basis for LiDAR–camera fusion and consistent embodied perception.
    }
    \label{fig:lidar_cam_calib}
    \vspace{-4mm}
\end{figure*}

\textbf{SP2.2) LiDAR-camera extrinsic calibration.}
To establish the rigid transformation between the LiDAR and the camera, we extend target-based calibration frameworks~\cite{beltran2022, zheng2025fast} to accommodate the unique sensing characteristics of solid-state LiDAR and ultra-wide-angle fisheye cameras. This results in a tailored calibration module, termed \textit{livox2cam}, specifically designed for our wrist-mounted multimodal perception setup.

Let ${}^L\mathbf T_C \in SE(3)$ denote the transformation from the LiDAR frame to the camera frame. As illustrated in Fig.~\ref{fig:lidar_cam_calib}, we employ a structured calibration target consisting of four circular holes and fiducial markers (Fig.~\ref{fig:lidar_cam_calib}B), which enables the extraction of geometrically consistent feature correspondences across sensing modalities. Specifically, the 3D positions of hole centers are independently estimated in both the LiDAR frame via edge-based geometric fitting (Fig.~\ref{fig:lidar_cam_calib}D) and the camera frame via marker-based pose recovery (Fig.~\ref{fig:lidar_cam_calib}C(i)). 

Given the resulting 3D--3D correspondences, the extrinsic calibration is formulated as a rigid alignment problem:
\begin{equation}
    \min_{{}^L\mathbf T_C} \sum_{i} \left\| \mathbf{p}_i^{C} - {}^L\mathbf T_C \cdot \mathbf{p}_i^{L} \right\|^2,
\end{equation}
where $\mathbf{p}_i^{L}$ and $\mathbf{p}_i^{C}$ denote corresponding hole centers in the LiDAR and camera coordinate frames, respectively. This formulation admits a closed-form solution via singular value decomposition (SVD), providing a fast and reliable initialization.

To further improve robustness under practical sensing conditions, the proposed \textit{livox2cam} module incorporates (i) scan-pattern-agnostic edge extraction for sparse and non-repetitive LiDAR measurements, (ii) ellipse-based fitting to compensate for spot-induced edge dilation, and (iii) multi-frame joint optimization for enhanced geometric consistency. These designs are critical for handling the irregular sampling of solid-state LiDAR (Fig.~\ref{fig:lidar_cam_calib}C(ii)) and the strong nonlinear distortion introduced by fisheye optics.

The estimated transformation ${}^L\mathbf T_C$ plays a central role in the UMI-3D system in two aspects. First, it enables the recovery of camera trajectories from LiDAR-centric SLAM by transforming LiDAR poses into the camera frame, thereby providing temporally consistent visual pose annotations. Second, it allows accurate projection of LiDAR points onto the image plane, forming the foundation for LiDAR–camera fusion, including depth association and patch-level 3D feature construction. Together, these calibration procedures ensure that all sensing modalities are geometrically aligned within a unified spatial representation centered at the end-effector (Fig.~\ref{fig:lidar_cam_calib}A), enabling robust embodied perception and interaction in real-world environments.

\begin{figure*}[ht]
    \centering
    \includegraphics[width=\linewidth]{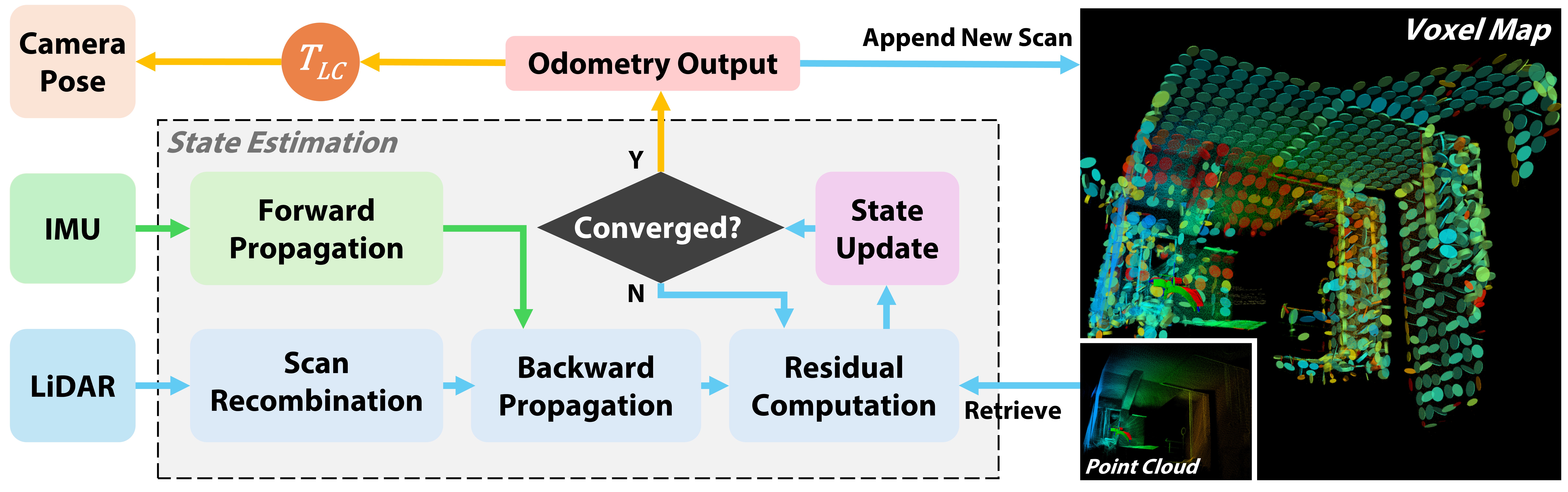}
    \caption{
    \textbf{Overview of the UMI-3D LiDAR–inertial odometry system.}
    The system follows an iterated error-state Kalman filtering (ESIKF) framework on differentiable manifolds.  IMU measurements drive high-frequency forward propagation, while LiDAR scans are processed through scan recombination and residual computation against a voxelized map representation.  The state is iteratively refined through backward propagation and update steps until convergence.  The estimated LiDAR trajectory is used to incrementally build a voxel-based map, and is further transformed to obtain the camera pose via the calibrated extrinsic transformation ${}^L\mathbf T_C$.  This tightly coupled pipeline enables robust state estimation and consistent geometric mapping under real-world conditions.
    }
    \label{fig:system_overview}
    \vspace{-4mm}
\end{figure*}

\textbf{SP3. LiDAR-Inertial Odometry}
\label{SP3}

\textbf{SP3.1) ESIKF on differentiable manifolds.}
To achieve consistent and drift-resistant state estimation under diverse real-world conditions, we adopt a LiDAR–inertial odometry framework based on an iterated error-state Kalman filter (ESIKF) on differentiable manifolds \cite{he2021kalman}, as illustrated in Fig.~\ref{fig:system_overview}. Taking the first IMU frame $I$ as the global frame $G$, the system state is defined on the manifold $\mathcal{M} = SO(3) \times \mathbb{R}^{15} \times SO(3) \times \mathbb{R}^{3}$ as
\begin{equation}
\mathbf x =
\begin{bmatrix}
^G{\mathbf R}_I^T & ^G\mathbf p_I^T & ^G\mathbf v_I^T &
\mathbf b_{\bm \omega}^T & \mathbf b_{\mathbf a}^T &
^G\mathbf g^T &
^{I}\mathbf R_{L}^T & ^{I}\mathbf p_{L}^T
\end{bmatrix}^T.
\end{equation}
The continuous-time kinematic model driven by IMU measurements is given by
\begin{equation}
\begin{aligned}
^G\dot{\mathbf R}_I&={}^G{\mathbf R}_I \lfloor \bm \omega_m -  \mathbf b_{\bm \omega} - \mathbf n_{\bm \omega} \rfloor_\wedge,\quad ^G\dot{\mathbf p}_I={}^G\mathbf v_I,\\
^G\dot{\mathbf v}_I&={}^G{\mathbf R}_I \left( \mathbf a_m - \mathbf b_{\mathbf a} - \mathbf n_{\mathbf a} \right) + {}^G\mathbf g,
\end{aligned}
\end{equation}
with IMU biases modeled as random walks. Using the $\boxplus$ operator, the system is discretized as
\begin{equation}
\mathbf x_{i+1} = \mathbf x_i \boxplus \left( \Delta t \ \mathbf f(\mathbf x_i, \mathbf u_i, \mathbf w_i) \right),
\end{equation}
which provides high-frequency motion prediction. The ESIKF iteratively refines the state on the manifold to handle the nonlinearity of the measurement model. For detailed derivations and complete symbol definitions, readers are referred to \cite{xu2022fast,he2021kalman}.

\textbf{SP3.2) LiDAR-centric Measurement Model.}
To correct the predicted state, we construct a LiDAR-centric measurement model based on voxelized probabilistic plane features. 
Given a LiDAR point ${}^L\mathbf p_i$, its global position is
\begin{equation}
{}^G\mathbf p_i = {}^G\mathbf R_I \left( {}^I\mathbf R_L \ {}^L\mathbf p_i + {}^I\mathbf p_L \right) + {}^G\mathbf p_I.
\end{equation}
Each voxel is represented as a plane $\Pi = (\mathbf n, \mathbf q)$ with covariance $\Sigma_{\Pi}$, capturing map uncertainty. The point-to-plane residual is defined as
\begin{equation}
r_i = \mathbf n^T \left( {}^G\mathbf p_i - \mathbf q \right),
\end{equation}
which is modeled as a Gaussian random variable
\begin{equation}
r_i \sim \mathcal{N}(0, \Sigma_{r_i}),
\end{equation}
where the residual covariance accounts for both point-wise and plane-wise uncertainties:
\begin{equation}
\Sigma_{r_i} =
\mathbf n^T \Sigma_{{}^G\mathbf p_i} \mathbf n
+
\mathbf J_{\Pi} \Sigma_{\Pi} \mathbf J_{\Pi}^T,
\end{equation}
with $\mathbf J_{\Pi} =
\begin{bmatrix}
({}^G\mathbf p_i - \mathbf q)^T & -\mathbf n^T
\end{bmatrix}$.
To ensure robust data association, we adopt a Mahalanobis distance gating:
\begin{equation}
r_i^T \Sigma_{r_i}^{-1} r_i < \tau.
\end{equation}
The measurement function is written as $\mathbf z_i = h(\mathbf x) + \mathbf v_i$, and linearized as
\begin{equation}
r_i \approx h(\hat{\mathbf x}) + \mathbf H_i \delta \mathbf x,
\end{equation}
which is iteratively solved within the ESIKF framework until convergence. 
This uncertainty-aware formulation enables adaptive residual weighting and improves robustness under sparse observations, long-range sensing, and geometric degeneracy, and supports consistent 3D map construction within the voxel-based representation shown in Fig.~\ref{fig:system_overview}. For detailed derivations, uncertainty propagation, and implementation details, readers are referred to \cite{yuan2022efficient}.

\textbf{SP3.3) Coordinate System Convention.}

We adopt a unified coordinate system convention to ensure consistent spatial-temporal alignment across all sensing and actuation modules. This formulation is particularly important in embodied manipulation settings, where perception, state estimation, and control must operate within a shared spatial reference.

Specifically, for each sequence, the first IMU frame $I_0$ is defined as the global reference frame $G$, and all subsequent states are estimated incrementally with respect to this frame.  The LiDAR-inertial odometry is formulated in the IMU-centric frame, where the transformation between the IMU and LiDAR frames, denoted as ${}^I\mathbf T_L$, is fixed and provided by factory calibration.  The extrinsic transformation between the LiDAR and camera frames, ${}^L\mathbf T_C$, is obtained through the calibration procedure described in Sec.~SP2.2, and the transformation between the LiDAR and the tool center point (TCP), ${}^L\mathbf T_{TCP}$, is determined by the rigid mechanical design of the end-effector.

\begin{figure}[!t]
    \centering
    \includegraphics[width=0.95\linewidth]{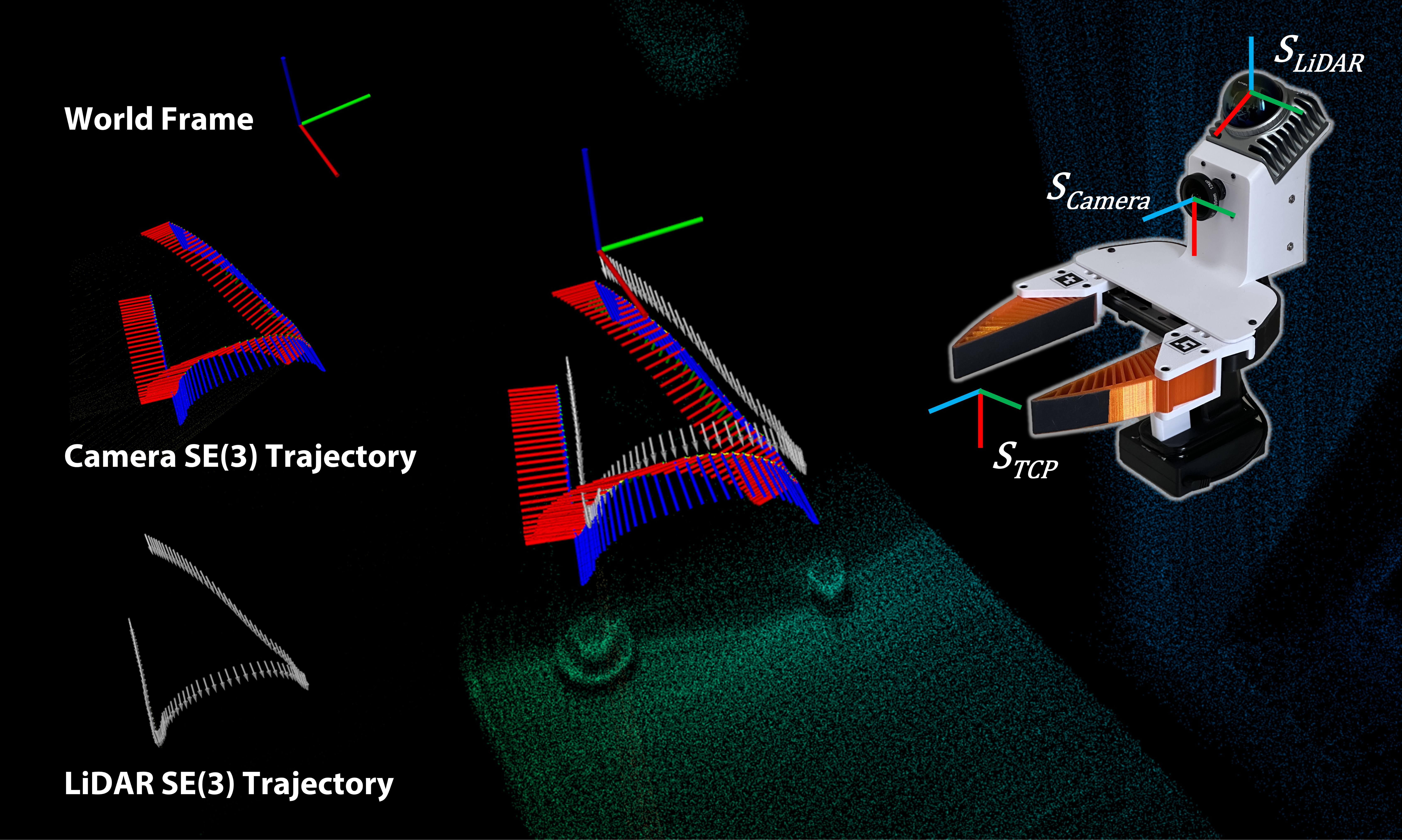}
    \caption{
    \textbf{Unified coordinate system and frame transformations in UMI-3D.}
    The global frame $G$ is initialized as the first IMU frame, and all states are estimated incrementally with respect to this reference. 
    The LiDAR trajectory is directly estimated through LiDAR–inertial odometry, while the camera trajectory is obtained by transforming LiDAR poses using the calibrated extrinsic ${}^L\mathbf T_C$. 
    The transformation between the LiDAR and IMU frames ${}^I\mathbf T_L$ is fixed and provided by factory calibration, and the transformation between the LiDAR and the tool center point (TCP), ${}^L\mathbf T_{TCP}$, is defined by the mechanical design of the end-effector. 
    This unified formulation enables consistent representation and transformation across perception and manipulation frames.
    }
    \vspace{-5mm}
    \label{fig:coordinate}
\end{figure}

These transformations define a consistent chain that connects perception and action spaces.  As illustrated in Fig.~\ref{fig:coordinate}, LiDAR poses estimated from odometry serve as the central reference, from which camera trajectories and manipulation frames are derived through fixed extrinsic transformations.  This unified formulation enables seamless transformation across camera, LiDAR, IMU, and manipulation frames, ensuring coherent perception, state estimation, and control within a common spatial reference.

\begin{figure*}[t]
    \centering
    \includegraphics[width=\linewidth]{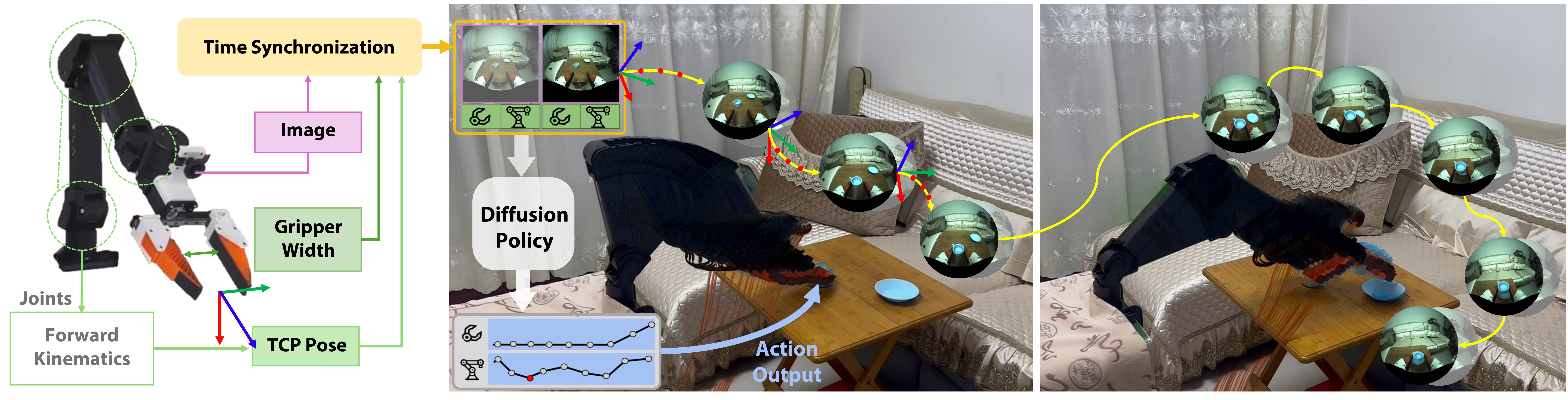}
    \caption{
    \textbf{Policy interface and relative action representation in UMI-3D.}
    (\textbf{Left}) The policy takes synchronized multimodal observations, including RGB images, relative end-effector (EE) poses, and gripper states, with explicit latency alignment across sensing and actuation streams. A diffusion policy predicts a sequence of future EE poses, which are executed in a receding-horizon manner with temporal ensembling. 
    (\textbf{Right}) Actions are represented as relative $SE(3)$ trajectories with respect to the current EE frame, illustrated by overlaid multi-step motions. 
    }
    \label{fig:method}
    \vspace{-4mm}
\end{figure*}

\textbf{SP4. Data Representation and Packaging for Policy Learning}
\label{SP4}

To bridge geometric perception with policy learning, we design a multimodal data representation and packaging pipeline that transforms raw multi-sensor recordings into temporally aligned and learning-ready trajectories.  Rather than treating perception and action as separate streams, this design explicitly couples visual observations, geometric structure, and motion into a unified representation, enabling consistent policy learning under embodied settings.

The pipeline takes synchronized LiDAR–camera trajectories and RGB observations as input, and produces a compressed Zarr-based replay buffer for downstream diffusion policy training.

\textbf{SP4.1) Multi-sensor synchronization and video alignment.}
Starting from raw ROS bag recordings, we first reorganize the data into a temporally aligned structure.  Camera frames are extracted and converted into MP4 videos while preserving per-frame timestamps. To ensure accurate synchronization with LiDAR scans, we adopt a strict alignment strategy where each LiDAR frame is associated with its nearest camera observations within a temporal gating window, enforcing a consistent cross-modal sampling ratio. The aligned data are then stored in a unified directory structure, where each sequence contains synchronized visual streams and timestamp annotations.  This process establishes a common temporal reference across all sensing modalities.

\textbf{SP4.2) LiDAR-inertial trajectory association.}
For each aligned sequence, we run the proposed LiDAR-inertial odometry to estimate the sensor trajectory in the global frame.  The SLAM system outputs a time-stamped trajectory file (\textit{camera\_trajectory.csv}), which is temporally consistent with the extracted video frames.  A batch processing pipeline automatically launches SLAM, monitors its execution, and associates the resulting trajectory with each demonstration sequence.  This step establishes a dense correspondence between visual observations and their underlying 3D motion.

\textbf{SP4.3) Frame-level alignment and state extraction.}
To construct training-ready trajectories, we further align all data streams at the frame level.  Using the video timestamps as the primary timeline, we associate SLAM poses and auxiliary signals via nearest-neighbor matching with a tight temporal tolerance (e.g., 10 ms).  In addition, fiducial markers (ArUco tags) are detected to estimate object and gripper states, enabling the extraction of manipulation-relevant signals such as end-effector pose and gripper width. All aligned signals are organized into episodic segments, forming structured trajectories for policy learning.

\textbf{SP4.4) Replay buffer construction.}
Finally, we convert the aligned dataset into a Zarr-based replay buffer for efficient training.  Each episode contains synchronized multimodal observations, including RGB images, end-effector poses, rotations, and gripper states.  Images are optionally rectified and compressed, while low-dimensional states are stored as temporally consistent sequences. The resulting dataset supports efficient random access and parallel loading, making it well-suited for large-scale diffusion policy training. Overall, this pipeline defines a unified multimodal representation that tightly couples perception and action, enabling scalable and robust learning of visuomotor policies from LiDAR-enhanced demonstrations.

\begin{figure*}[t] 
    \centering
    \includegraphics[width=\linewidth]{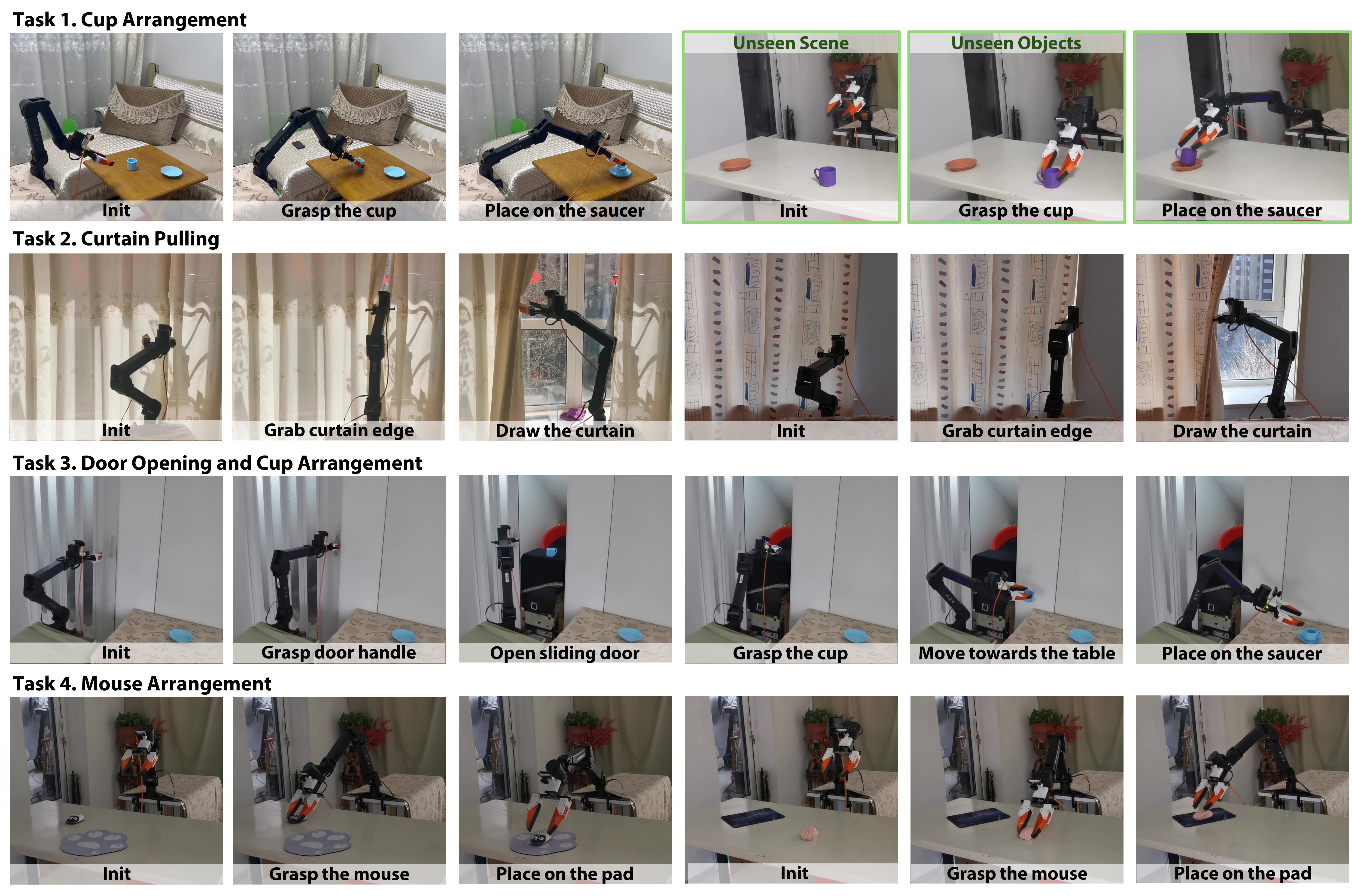}
    \caption{
    \textbf{Evaluation tasks and policy rollouts.}
    We evaluate UMI-3D across four real-world manipulation tasks. 
    \textbf{Task 1. Cup arrangement}, including both in-distribution and unseen scene/object generalization. 
    \textbf{Task 2. Curtain pulling}, involving large deformable objects under varying lighting conditions. 
    \textbf{Task 3. Door opening and cup arrangement}, a long-horizon task requiring interaction with articulated structures followed by object manipulation. 
    \textbf{Task 4. Mouse arrangement}, used to evaluate cross-embodiment policy transfer without retraining. Each row shows representative rollout sequences illustrating key interaction stages. Please check videos on our \href{https://umi-3d.github.io}{website} for more details.
    }
    \label{fig:task}
    \vspace{-4mm}
\end{figure*}

\subsection{Policy Learning}
\label{sec:policy_learning}

We adopt a diffusion policy framework to learn visuomotor control from demonstration data.  The design largely follows the UMI framework \cite{chi2024universal}, with modifications to account for LiDAR-enhanced data and a low-latency sensing pipeline, as illustrated in Fig.~\ref{fig:method}, which summarizes the multimodal observation design, latency-aware policy interface, and the relative action representation.

\textbf{PD1. Observation and action latency alignment.}
\label{PD1}

\textbf{PD1.1) Low-latency visual sensing.}
Compared to UMI, which relies on a GoPro-based pipeline, we employ an industrial CMOS camera with a direct USB interface.  This design significantly reduces sensing and transmission latency, providing more accurate temporal alignment between observations and actions.

\textbf{PD1.2) Action latency matching.}
The robotic arm and gripper are controlled via a unified CAN bus, where the gripper is driven by the same motor control system as the arm joints.  This ensures consistent actuation latency across all degrees of freedom, improving synchronization between perception and control.

\textbf{PD2. Relative end-effector pose representation.}
\label{PD2}

End-effector (EE) pose is central to both observation and action spaces.  We represent all EE poses relative to the current end-effector frame to avoid dependence on embodiment-specific coordinates.

\textbf{PD2.1) Relative EE trajectory as action representation.}
\label{PD2.1}

Action sequences are represented as relative $SE(3)$ transformations with respect to the current EE pose.  This representation improves robustness to tracking errors and camera displacement.

\textbf{PD2.2) Relative EE trajectory as proprioception.}
\label{PD2.2}

Historical EE poses are represented as a relative trajectory. With a short observation horizon, this formulation implicitly provides velocity information to the policy.  This design is consistent with the incremental state estimation used in the SLAM pipeline \hyperref[SP3]{SP3}, enabling a unified representation across perception and control.

\textbf{PD3. Policy training and deployment.}
\label{PD3}

We train a diffusion policy  \cite{chi2025diffusion} on the constructed replay buffer.  The policy takes as input RGB observations and proprioceptive states, and predicts a sequence of future EE poses.  During deployment, predicted trajectories are executed using a receding horizon control strategy with temporal ensembling for improved stability.

Overall, the policy design maintains the simplicity and generality of the UMI framework, while benefiting from improved data quality enabled by LiDAR-based state estimation and a low-latency sensing pipeline.

\begin{figure*}[!t]
\centering
\includegraphics[width=0.9\textwidth]{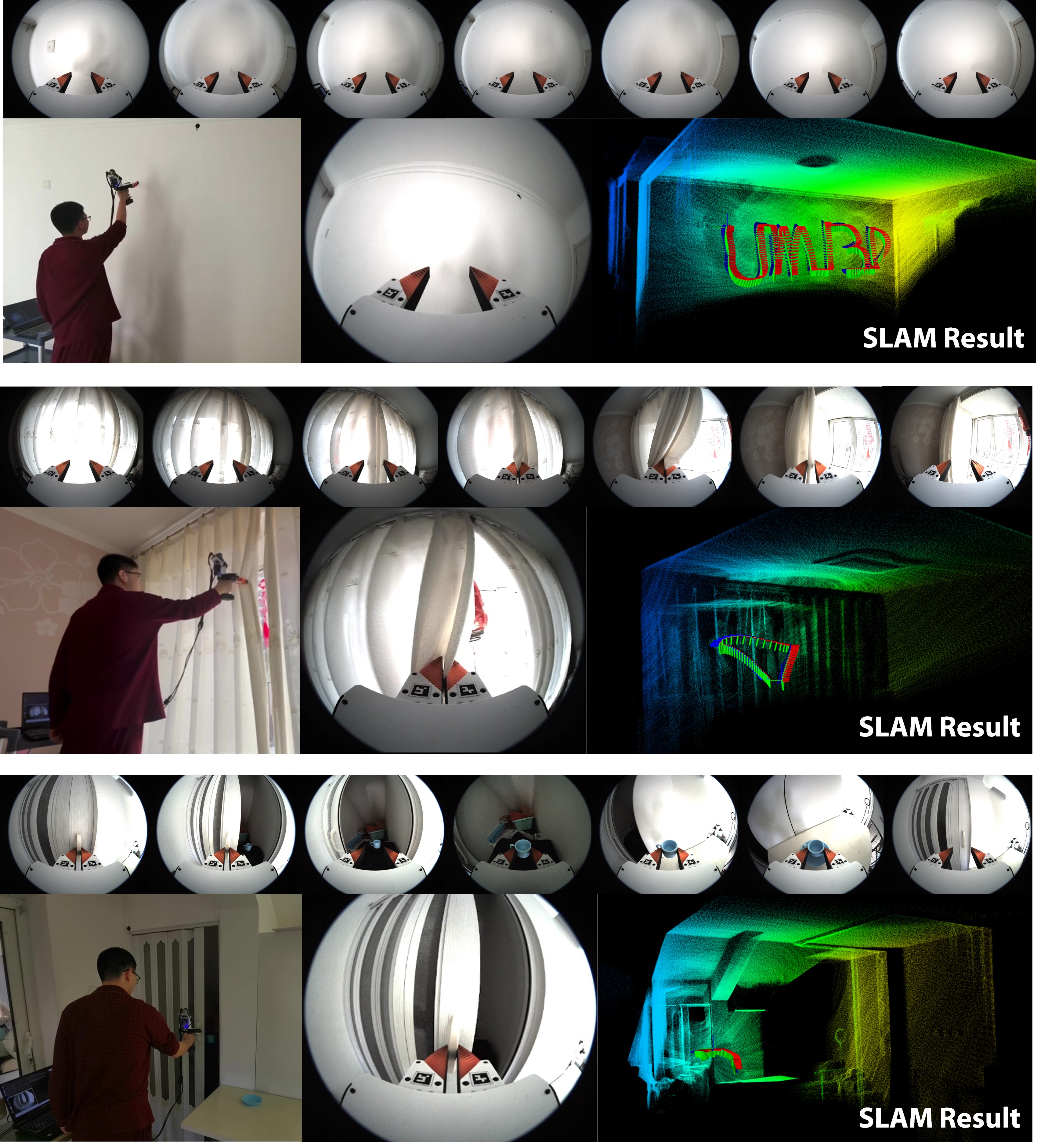}
\caption{
\textbf{Robust LiDAR-centric SLAM under challenging real-world conditions.} 
Three representative scenarios are shown: 
(1) a textureless white wall with minimal visual features, 
(2) rapid curtain pulling under strong illumination changes with large deformable motion, and 
(3) a long-horizon manipulation task involving articulated structures and dynamic occlusions.  
In all cases, UMI-3D achieves stable and accurate pose estimation and mapping, 
despite conditions that are highly challenging for vision-only SLAM. Please check videos on our \href{https://umi-3d.github.io}{website} for more details.
}
\label{fig:slam_result}
\end{figure*}

\begin{figure*}
\centering
\includegraphics[width=\linewidth]{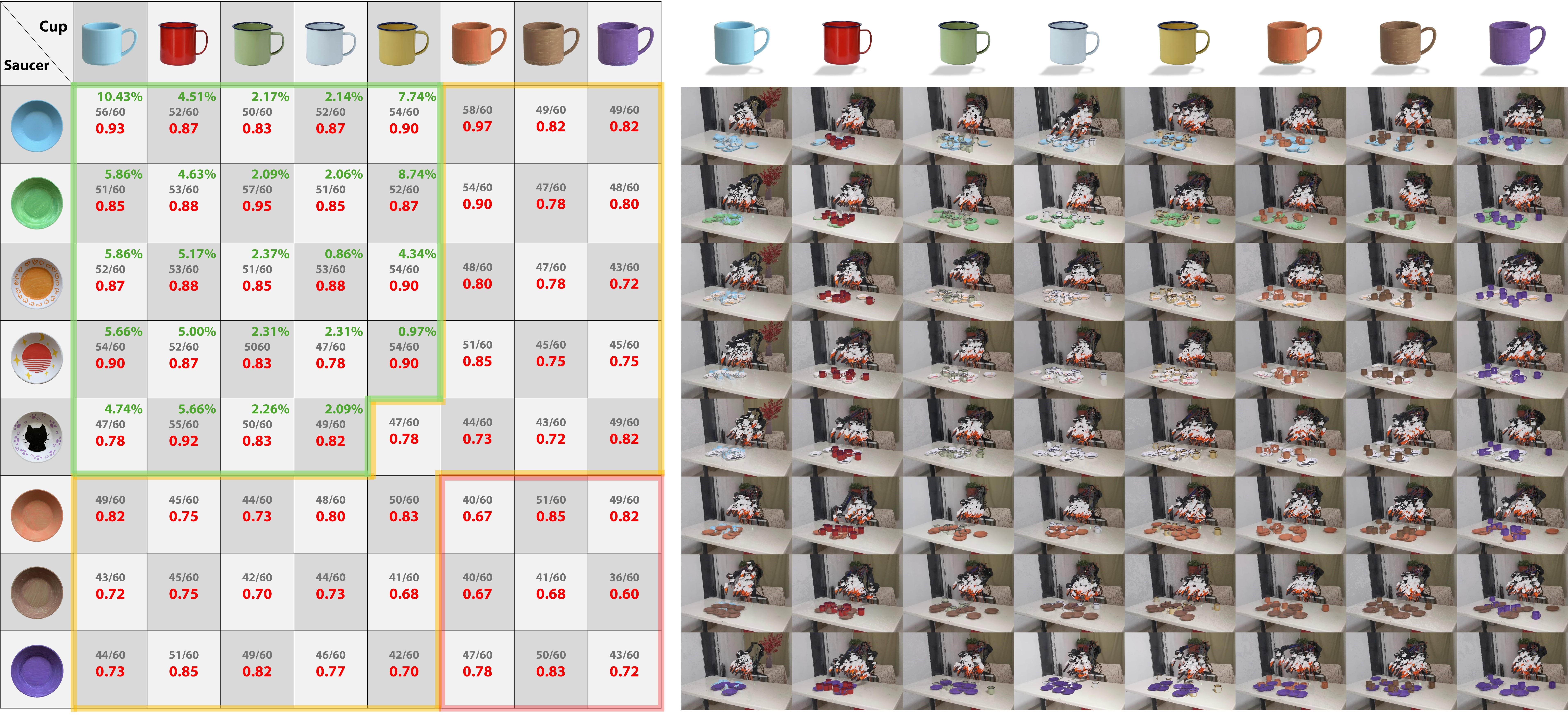}
\caption{
\textbf{Cup arrangement performance across object combinations.}
\textbf{Left:} Quantitative results over 8 $\times$ 8 cup–saucer combinations. 
Each cell corresponds to one object pair evaluated over 10 trials. 
The number in the top-right corner indicates the percentage of this combination in the 3,500 training demonstrations. 
The gray value denotes the accumulated score (out of 60), and the red value shows the normalized score. 
Green boxes denote \textcolor[rgb]{0.2,0.6,0.2}{seen combinations} that appear in the training data. 
Orange boxes represent \textcolor[rgb]{0.85,0.5,0.1}{partially unseen combinations}, where either the cup or the saucer is unseen. 
Red boxes correspond to \textcolor[rgb]{0.8,0.2,0.2}{fully unseen combinations}. 
The average normalized scores for seen, partially unseen, and fully unseen regions are \textcolor[rgb]{0.2,0.6,0.2}{0.863}, \textcolor[rgb]{0.85,0.5,0.1}{0.788}, and \textcolor[rgb]{0.8,0.2,0.2}{0.736}, respectively, indicating graceful performance degradation under distribution shift.
\textbf{Right:} Visualization of the 10 trial initializations for each object combination. 
Each row corresponds to a fixed saucer, and each column corresponds to a cup instance. 
The overlaid images illustrate spatial variations in initial configurations, demonstrating consistent performance across diverse setups. 
Please check our \href{https://umi-3d.github.io}{website} for more comparison videos. 
\vspace{-4mm}
}
\label{fig:cup_table}
\end{figure*}

\section{Evaluations}

We evaluate the proposed UMI-3D system from three complementary aspects:

\begin{itemize}
    \item \textbf{SLAM robustness and data quality:} 
    To what extent does LiDAR-centric sensing improve the robustness of pose estimation under challenging real-world conditions, including textureless surfaces, large dynamic objects, and severe occlusions? How does this improvement translate into higher-quality demonstration data?

    \item \textbf{Capability, generalization, and task coverage:} 
    Can UMI-3D enable effective policy learning for manipulation tasks ranging from standard pick-and-place to interactions that are challenging or infeasible under the vision-only UMI configuration.  Furthermore, do policies trained with UMI-3D data generalize to unseen environments and objects?

    \item \textbf{Cross-embodiment compatibility:} 
    Can existing UMI-trained policies be directly deployed on UMI-3D hardware without retraining?
\end{itemize}

To assess capability and generalization, we evaluate UMI-3D on four real-world robotic manipulation tasks across both narrow-domain and unseen environments, as illustrated in Fig.~\ref{fig:task}. These tasks are designed to progressively evaluate system capability, starting from basic manipulation and generalization (Task 1), and extending to interaction scenarios that are challenging for vision-only UMI, including large deformable objects (Task 2) and large articulated objects (Task 3).  Finally, we assess cross-platform compatibility by deploying policies trained with the original UMI system on UMI-3D hardware without modification (Task 4).

\section{Capability Experiments}

\mypara{Robust SLAM for Reliable Data Collection}
Before evaluating policy performance, we first demonstrate the robustness of the LiDAR-centric SLAM system in UMI-3D, which enables accurate and consistent data collection in real-world environments. As shown in Fig.~\ref{fig:slam_result}, we evaluate SLAM performance in three representative challenging scenarios: a textureless white wall, a deformable curtain under strong lighting variation, and a long-horizon manipulation task involving articulated objects and dynamic occlusions.  These conditions are known to be highly challenging for vision-based SLAM due to lack of texture, large non-rigid motion, and drastic illumination changes. Across all scenarios, UMI-3D maintains stable and drift-resistant pose estimation while producing consistent 3D map reconstruction. 

Instead of focusing on peak localization accuracy, we prioritize consistent and reliable SLAM performance, ensuring uniformly high accuracy across diverse and challenging real-world settings. In addition, the LiDAR-based SLAM system produces metrically consistent 3D maps, which provide two practical benefits. First, the reconstructed geometry enables intuitive quality inspection, where artifacts such as wall thickness, structural continuity, and layering directly reflect pose estimation accuracy.  Second, the global map serves as a spatial reference for aligning data across multiple sequences and devices, facilitating scalable data collection. Overall, this robustness ensures the collection of well-aligned image–action pairs, which is critical for learning reliable manipulation policies.

\mypara{Task Evaluation Overview}
Building upon the reliable and geometrically consistent data enabled by LiDAR-centric SLAM, we evaluate how this improved data quality translates into downstream policy capability.  Specifically, we design a set of manipulation tasks that probe complementary aspects of embodied intelligence, including object generalization, deformable interaction, articulated manipulation, and cross-embodiment transfer.  These tasks are not only benchmarks of performance, but also serve as diagnostic tools to reveal the strengths and limitations of learning from LiDAR-enhanced demonstrations.

\vspace{-2mm}
\subsection{\textbf{Cup Arrangement}} 
\label{sec:eval_cup}

\mypara{Task}
The robot is required to grasp a cup and place it onto a saucer, as shown in Fig.~\ref{fig:task}. 
This task serves as a standard pick-and-place benchmark for evaluating both manipulation capability and generalization performance.

\mypara{Experimental Setup}
We collect 3,500 demonstration trajectories using UMI-3D and train a diffusion policy with a CLIP \cite{radford2021learning} ViT-L/14 encoder \cite{dosovitskiy2020image}. 
Evaluation is conducted entirely in unseen environments using 8 cups and 8 saucers, resulting in 64 object combinations. 
Each combination is evaluated over 10 trials, yielding a total of 640 evaluation runs spanning seen, partially unseen, and fully unseen object configurations.

\mypara{Evaluation Metric}
Each trial is scored using a two-stage protocol, covering grasping and placement. 
The grasping stage evaluates contact and grasp success, while the placement stage evaluates the stability and accuracy of placing the cup onto the saucer. 
Each stage is scored from 0 to 3, resulting in a maximum score of 6 per trial. 
We report normalized scores for all results. 
Detailed scoring criteria are provided in Sec.~\ref{sec:cup_scoring}.

\mypara{Generalization Performance}
As shown in Fig.~\ref{fig:cup_table}, the policy achieves strong performance across all object combinations. 
For seen object pairs, the average normalized score reaches 0.863. 
Under partial distribution shift (either cup or saucer unseen), performance decreases moderately to 0.788, and further to 0.736 in fully unseen scenarios. 
Despite this gradual degradation, performance remains consistently high, indicating that the learned policy generalizes beyond memorizing specific object instances.

\mypara{Failure Analysis}
A notable failure mode is observed for the brown saucer (seventh row in Fig.~\ref{fig:cup_table}), which consistently yields the lowest scores. 
In many cases, the policy successfully grasps the cup and reaches the correct placement region, but hesitates to release the object or places it outside the saucer.

We attribute this behavior to visual ambiguity. 
The brown saucer exhibits low contrast with the surrounding environment and deviates from the training data distribution, making it difficult for the policy to reliably localize the placement target. 
This leads to uncertainty in fine-grained placement decisions, even when the overall task structure is correctly executed.

\mypara{Discussion}
These results demonstrate that UMI-3D enables robust manipulation and strong generalization across object variations and unseen environments. 
At the same time, the observed failure cases highlight that precise placement remains fundamentally limited by perceptual quality, motivating the need for more geometry-aware or multimodal representations in challenging scenarios.

\begin{figure}
\centering
\includegraphics[width=\linewidth]{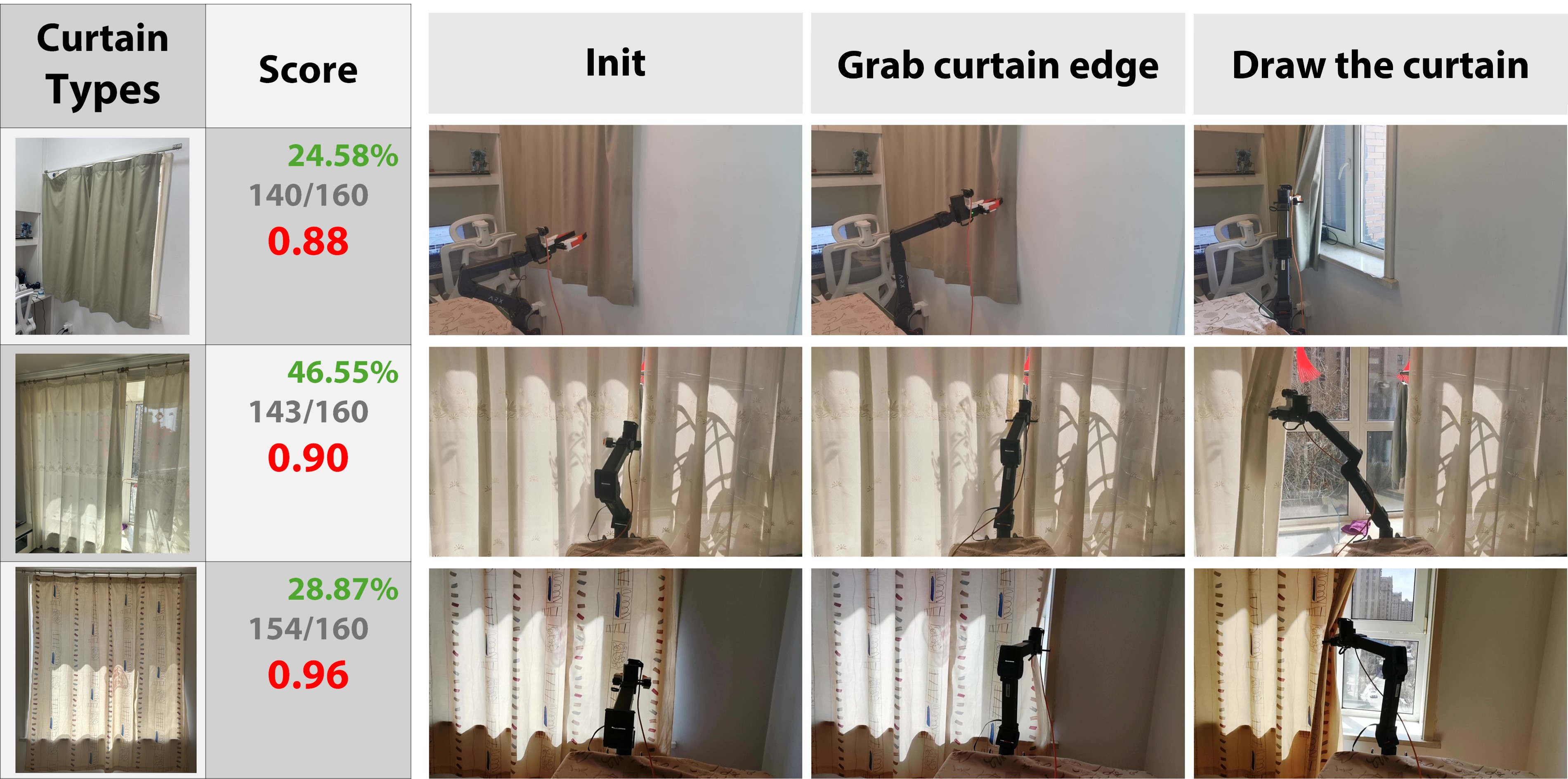}
\caption{
\textbf{Curtain pulling performance under varying conditions.}
\textbf{Left:} Three representative curtain types used in training and evaluation, along with their occurrence ratios in the 769 demonstration trajectories. For each curtain type, we report the accumulated score (out of 160) and the normalized score across 40 trials. 
\textbf{Right:} Representative execution sequences for each curtain type, including the initial configuration, grasping of the curtain edge, and pulling motion. The examples illustrate robustness to different material properties, textures, and challenging lighting conditions such as strong illumination and backlighting. 
Despite relying only on visual inputs at inference time, the learned policy achieves consistent performance, enabled by high-quality data collected with LiDAR-based pose estimation. Please check our \href{https://umi-3d.github.io}{website} for more comparison videos. 
\vspace{-4mm}
}
\label{fig:curtain}
\end{figure}

\begin{figure*}
\centering
\includegraphics[width=\linewidth]{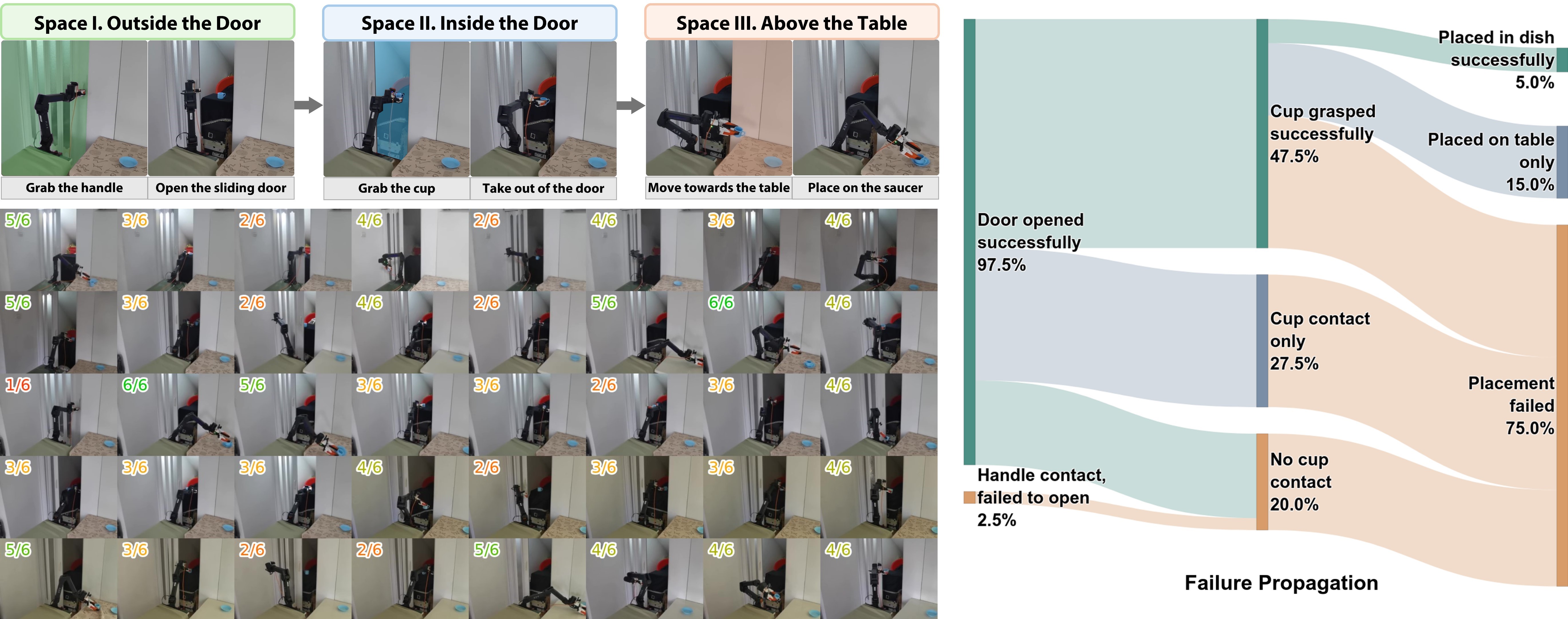}
\caption{\textbf{Long-horizon manipulation: door opening, cup grasping, and placement.} 
\textbf{Left:} Task decomposition into three spatial regions: outside the door, inside the cabinet, and above the table. The robot sequentially performs door opening, cup retrieval, and placement, with all trials and corresponding scores shown.  Please check our \href{https://umi-3d.github.io}{website} for more comparison videos.
\textbf{Right:} Failure propagation analysis using a Sankey diagram. While door opening achieves a high success rate (97.5\%), errors accumulate in subsequent stages, with reduced success in cup grasping (47.5\%) and final placement (5.0\%). The diagram illustrates how early-stage failures propagate and prevent downstream success, highlighting the challenges of long-horizon sequential manipulation.
\vspace{-4mm}
}
\label{fig:door}
\end{figure*}

\subsection{\textbf{Curtain Pulling}}
\label{sec:eval_curtain}

\mypara{Task}
The robot is required to grasp the edge of a curtain and pull it open, as illustrated in Fig.~\ref{fig:curtain}.  This task involves manipulation of a large deformable object and requires accurate interaction with thin, low-texture structures under varying lighting conditions.

\mypara{Experimental Setup}
We collect 769 demonstration trajectories using UMI-3D and train a diffusion policy with a DINOv2 ViT-L/14  \cite{oquab2023dinov2} encoder.  The training set includes three curtain types with different materials and appearances.  Evaluation is performed under diverse lighting conditions, including strong illumination and backlit scenarios, resulting in a total of 120 trials.

\mypara{Evaluation Metric}
Each trial is evaluated using a two-stage scoring protocol, including grasping and pulling. The grasping stage evaluates whether the curtain edge is successfully contacted and grasped, while the pulling stage measures the extent of curtain displacement. Each stage is scored from 0 to 2, resulting in a maximum score of 4 per trial. We report normalized scores for all results. Detailed scoring criteria are provided in Sec.~\ref{sec:curtain_scoring}.

\mypara{Performance}
As shown in Fig.~\ref{fig:curtain}, the policy achieves strong performance across all curtain types, with normalized scores of 0.88, 0.90, and 0.96, respectively. The system remains robust under significant variations in lighting and appearance, demonstrating reliable grasping and pulling behaviors for deformable objects.

\mypara{Failure Analysis}
Failure cases mainly occur under strong backlighting or low contrast conditions, where the curtain edge becomes difficult to distinguish from the background.  In such scenarios, the policy may fail to establish a stable grasp or may grasp at suboptimal locations, leading to incomplete pulling.

\mypara{Discussion}
These results highlight a key advantage of UMI-3D in handling deformable object manipulation under challenging visual conditions.  The benefit of LiDAR lies in enabling accurate and drift-resistant pose estimation during data collection, resulting in better-aligned image–action pairs for policy learning. In this task, large portions of the wrist-mounted camera view are dominated by dynamic, low-texture curtain motion under strong illumination changes, which makes pose estimation impossible for vision-only SLAM.  This makes it possible to learn reliable policies for tasks that are difficult to collect using vision-only SLAM, while the learned policy itself operates purely on visual inputs during both training and deployment.

\subsection{\textbf{Door Opening and Cup Placement}}
\label{sec:eval_door}

\mypara{Task}
The robot is required to sequentially open a sliding door, grasp a cup inside, and place it onto a saucer on a nearby table, as illustrated in Fig.~\ref{fig:door}. 
Compared to curtain pulling, which involves coarse manipulation of deformable objects, this task requires fine-grained interaction with a large articulated object, followed by object transfer and precise placement. 
The three stages are strongly coupled both temporally and spatially, forming a long-horizon manipulation task.

\mypara{Experimental Setup}
We collect 340 demonstration trajectories using UMI-3D and train a diffusion policy with a DINOv2 ViT-L/14  \cite{oquab2023dinov2} encoder. 
Evaluation is conducted in the same environment as training, with randomized initial states, resulting in 40 trials.

\mypara{Evaluation Metric}
Each trial is evaluated using a three-stage scoring protocol, including door opening, cup grasping, and placement.  Each stage is scored from 0 to 2, resulting in a maximum score of 6 per trial.  We report normalized scores for all results. Detailed scoring criteria are provided in Sec.~\ref{sec:door_scoring}.

\mypara{Performance and Failure Propagation}
As shown in Fig.~\ref{fig:door}, the door opening stage achieves a high success rate of 97.5\%, indicating that the policy reliably learns articulated object interaction. However, performance decreases in subsequent stages, with cup grasping success at 47.5\% and final placement success at only 5.0\%. The Sankey diagram reveals a clear failure propagation pattern, where errors accumulate across stages and early-stage failures prevent downstream success.

\mypara{Failure Analysis}
We identify two primary sources of failure. First, a significant portion (32.5\%) of trials successfully complete door opening and cup grasping but fail during placement due to violations of the robot's inverse kinematics constraints. This suggests a mismatch between the demonstrated motions and the robot's feasible configuration space, indicating the need for stricter kinematic filtering or embodiment-aware policy constraints. Second, the relatively low grasping success rate is attributed to limited and less diverse training data. Compared to the cup arrangement task, which benefits from large-scale and diverse demonstrations, the smaller dataset in this task leads to reduced robustness in grasping performance.

\mypara{Discussion}
These results highlight the challenges of long-horizon manipulation, where errors compound across sequential stages. While UMI-3D enables reliable data collection for complex interactions such as door opening, successful execution of multi-stage tasks further requires consistency between data distribution, robot embodiment, and control feasibility. This suggests that beyond improving data quality, incorporating kinematic constraints and increasing data diversity are critical for scaling to more complex sequential tasks.

\begin{figure*}[t]
\centering
\includegraphics[width=\linewidth]{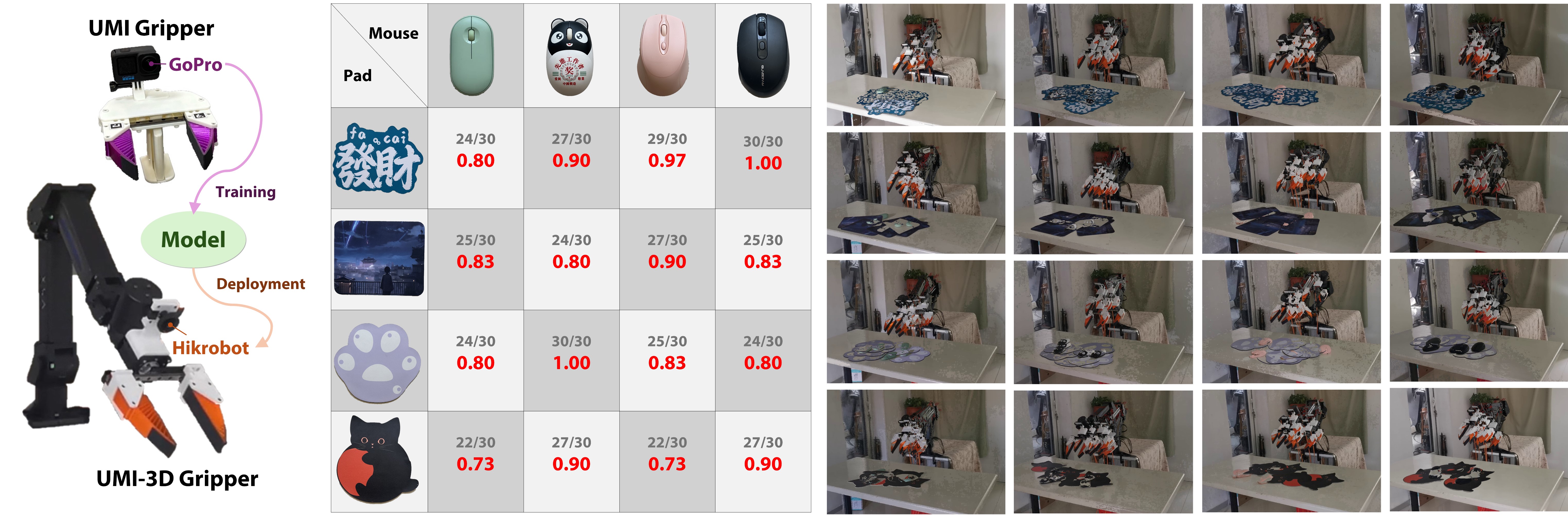}

\caption{
\textbf{Cross-embodiment policy transfer from UMI to UMI-3D.} 
\textbf{Left:} Illustration of the training–deployment pipeline. 
Policies are trained using the original UMI system and directly deployed on UMI-3D hardware without finetuning.
\textbf{Middle:} Quantitative results across 4 $\times$ 4 mouse–pad combinations in unseen environments. 
Each cell reports the accumulated score (out of 30) and normalized score over 5 trials.
\textbf{Right:} Representative execution sequences, including grasping and placement. 
The results demonstrate that policies trained under the original UMI setup generalize effectively to UMI-3D, despite changes in hardware and sensing configuration.
\vspace{-4mm}
}
\label{fig:mouse}
\end{figure*}

\subsection{\textbf{Cross-Embodiment Transfer: Mouse Placement}}
\label{sec:eval_cross}

\mypara{Task}
The robot is required to grasp a mouse and place it onto a mouse pad, as illustrated in Fig.~\ref{fig:mouse}.  This task evaluates whether policies trained using the original UMI system can be directly deployed on UMI-3D hardware without retraining, despite changes in sensing and embodiment.

\mypara{Experimental Setup}
We use a pretrained policy from prior work on data scaling laws in imitation learning~\cite{hu2024data}, trained entirely using the original UMI system. 
Without any finetuning, the policy is directly deployed on UMI-3D hardware. 
Evaluation is conducted in unseen environments with four mouse instances and four mouse pads, resulting in 16 object combinations. 
Each combination is evaluated over 5 trials, yielding a total of 80 trials.

\mypara{Evaluation Metric}
Each trial is evaluated using a two-stage scoring protocol, including grasping and placement. 
Each stage is scored from 0 to 3, resulting in a maximum score of 6 per trial. 
We report normalized scores for all results. 
Detailed scoring criteria are provided above.

\mypara{Cross-Embodiment Performance}
As shown in Fig.~\ref{fig:mouse}, the policy achieves consistently strong performance across all object combinations, with normalized scores ranging from 0.73 to 1.00. 
Despite differences in hardware configuration and sensing modality, the policy maintains stable grasping and placement behaviors in unseen environments.

\mypara{Discussion}
These results demonstrate that UMI-3D preserves compatibility with policies trained using the original UMI system.  This indicates that UMI-3D maintains a closely aligned visual observation space with the original UMI setup, despite differences in hardware design, enabling seamless cross-embodiment policy transfer.  It also suggests that the visual observations collected by UMI-3D are largely compatible with those from the original system, potentially enabling joint training across datasets to further improve policy performance through data scaling.

\section{Limitations and Future Work}

Despite the promising results, UMI-3D still has several limitations that suggest important directions for future research.

\textbf{Hardware design and ergonomics.}
The current UMI-3D hardware introduces additional weight due to the integration of LiDAR and supporting components, which can lead to user fatigue during prolonged data collection.  Future work could explore lightweight materials, more compact sensor integration, and improved gripper and handle design to enhance usability and ergonomics for long-duration operation.

\textbf{Single-arm system limitation.}
UMI-3D currently focuses on a single-arm configuration, which limits its ability to perform tasks that require bimanual coordination or complex object stabilization.  Extending the system to a dual-arm setup would significantly expand the range of achievable manipulation tasks and enable richer forms of interaction.

\textbf{Limited utilization of 3D perception in policy learning.}
While LiDAR plays a critical role in improving SLAM robustness and data quality, the learned policies in this work rely primarily on visual observations at inference time.  However, the system inherently captures synchronized 3D geometric information during demonstrations, including object structure, spatial relationships, and interaction dynamics.  An important future direction is to incorporate such 3D information directly into policy learning, potentially enabling more robust and geometry-aware manipulation.

\textbf{Extension to mobile manipulation.}
The high-fidelity data collection capability of UMI-3D provides a strong foundation for extending from fixed-base manipulation to mobile manipulation scenarios.  Future work could integrate locomotion and manipulation into a unified data-to-policy pipeline, enabling robots to operate in larger and less structured environments and significantly expanding the scope of embodied intelligence.

Overall, addressing these limitations will further enhance the scalability, usability, and generality of UMI-3D, and help bridge the gap between data collection, perception, and embodied decision-making.

\section*{Acknowledgments}

The author would like to thank Prof. Fu Zhang and Dr. Zhengrong Xue for their insightful and constructive discussions. The author also gratefully acknowledges Prof. Yanyong Zhang for providing essential computational resources that enabled this work.


\bibliographystyle{plainnat}
\bibliography{references}

\newpage
\appendix
\renewcommand{\thesection}{A.\arabic{section}}
\renewcommand{\thefigure}{A\arabic{figure}}
\renewcommand{\thetable}{A\arabic{table}}
\setcounter{section}{0}
\setcounter{figure}{0}
\setcounter{table}{0}

\subsection{\textbf{Task Scoring Protocol}}
\label{sec:supp_scoring}

To quantitatively evaluate manipulation performance, we adopt structured, task-specific scoring protocols across all experiments. Each task is decomposed into multiple stages corresponding to key sub-goals, and each stage is independently scored. This design enables fine-grained analysis of system behavior and allows us to distinguish between different types of failure modes, such as grasping errors and placement inaccuracies. All scoring criteria follow a monotonic scale, where higher scores indicate more successful task execution.  This unified design ensures consistent evaluation across tasks with varying levels of complexity.

The total score for each trial is computed as the sum of scores across all stages:
\[
S_{\text{total}} = \sum_{k} S_k,
\]
where $S_k$ denotes the score of the $k$-th stage.

For each task, we report both the accumulated score over multiple trials and a normalized score defined as:
\[
S_{\text{norm}} = \frac{S_{\text{total}}}{S_{\text{max}}},
\]
where $S_{\text{max}}$ is the maximum achievable score.
\subsubsection{\textbf{Cup Arrangement}}
\label{sec:cup_scoring}

This task evaluates the robot’s ability to grasp a cup and place it onto a saucer.

\textbf{Grasping Stage}
\begin{itemize}
    \item \textbf{0: No contact.} The gripper does not touch the cup.
    \item \textbf{1: Partial contact.} The gripper touches the cup but fails to align for grasping.
    \item \textbf{2: Aligned but unsuccessful grasp.} Both inner sides of the gripper contact the cup, but lifting fails.
    \item \textbf{3: Successful grasp.} The cup is securely grasped and lifted.
\end{itemize}

\textbf{Placement Stage}
\begin{itemize}
    \item \textbf{0: Failed placement.} The cup is not placed on the saucer.
    \item \textbf{1: Unstable placement.} The cup contacts the saucer boundary or slides off after contact.
    \item \textbf{2: Partial placement.} The cup is placed on the saucer but lacks full support or is misaligned.
    \item \textbf{3: Stable placement.} The cup is fully and stably placed on the saucer.
\end{itemize}

\textbf{Total Score} The maximum score is 6 per trial.

\subsubsection{\textbf{Curtain Pulling}}
\label{sec:curtain_scoring}

This task evaluates manipulation of a deformable object through grasping and pulling.

\textbf{Grasping Stage}
\begin{itemize}
    \item \textbf{0: No contact.} The gripper does not touch the curtain.
    \item \textbf{1: Contact without grasp.} The gripper touches the curtain but fails to establish a stable grasp.
    \item \textbf{2: Successful grasp.} The curtain edge is successfully grasped.
\end{itemize}

\textbf{Pulling Stage}
\begin{itemize}
    \item \textbf{0: No movement.} The curtain is not pulled.
    \item \textbf{1: Partial pulling.} The curtain is partially opened.
    \item \textbf{2: Complete pulling.} The curtain is opened beyond 30 cm.
\end{itemize}

\textbf{Total Score} The maximum score is 4 per trial.

\subsubsection{\textbf{Door Opening and Cup Placement}}
\label{sec:door_scoring}

This task evaluates long-horizon sequential manipulation involving articulated object interaction and object transfer.

\textbf{Door Opening Stage}
\begin{itemize}
    \item \textbf{0: No contact.} The gripper does not touch the handle.
    \item \textbf{1: Contact without opening.} The gripper touches the handle but fails to open the door.
    \item \textbf{2: Successful opening.} The handle is grasped and the door is successfully opened.
\end{itemize}

\textbf{Cup Grasping Stage}
\begin{itemize}
    \item \textbf{0: No contact.} The gripper does not touch the cup.
    \item \textbf{1: Contact without grasp.} The gripper touches the cup but fails to grasp it.
    \item \textbf{2: Successful grasp.} The cup is securely grasped.
\end{itemize}

\textbf{Placement Stage}
\begin{itemize}
    \item \textbf{0: Failed placement.} The cup is not placed correctly.
    \item \textbf{1: Partial placement.} The cup is placed on the table or only partially contacts the saucer.
    \item \textbf{2: Successful placement.} The cup is fully placed onto the saucer.
\end{itemize}

\textbf{Total Score} The maximum score is 6 per trial.

\subsubsection{\textbf{Mouse Grasping and Placement}}
\label{sec:mouse_scoring}

This task evaluates fine-grained manipulation requiring precise grasping and controlled placement.

\textbf{Grasping Stage}
\begin{itemize}
    \item \textbf{0: No approach.} The gripper does not move toward the mouse or fails to establish contact.
    \item \textbf{1: Unstable contact.} The gripper reaches the mouse but fails to maintain a stable grasp.
    \item \textbf{2: Unstable grasp.} The mouse is grasped but slips or falls during lifting.
    \item \textbf{3: Stable grasp.} The mouse is successfully grasped without slippage.
\end{itemize}

\textbf{Placement Stage}
\begin{itemize}
    \item \textbf{0: Failed placement.} The mouse is not placed on the pad or is dropped from a height.
    \item \textbf{1: Incorrect placement.} The mouse is placed outside the pad or lands unstably.
    \item \textbf{2: Partial placement.} The mouse is partially placed or shifts due to unstable release.
    \item \textbf{3: Stable placement.} The mouse is fully and stably placed on the mouse pad.
\end{itemize}

\textbf{Total Score} The maximum score is 6 per trial.

\end{document}